\documentclass{article}

\PassOptionsToPackage{numbers, compress}{natbib}

\usepackage[preprint]{neurips_2025}

\usepackage{microtype}
\usepackage{graphicx}
\usepackage{booktabs} 
\usepackage{multirow}
\usepackage{xcolor}
\usepackage{tcolorbox}
\usepackage{colortbl}

\usepackage[utf8]{inputenc} 
\usepackage[T1]{fontenc}    
\usepackage{hyperref}       
\usepackage{url}            
\usepackage{booktabs}       
\usepackage{amsfonts}       
\usepackage{nicefrac}       
\usepackage{microtype}      
\usepackage{xcolor}         
\usepackage{tabularx} 
\usepackage{ragged2e}

\usepackage{amsmath}
\usepackage{amssymb}
\usepackage{mathtools}
\usepackage{amsthm}
\theoremstyle{definition} 
\usepackage{enumitem} 

\usepackage{amsfonts, mathrsfs}
\usepackage{makecell} 
\usepackage{multirow} 
\usepackage{wrapfig}
\usepackage[numbers]{natbib}
\usepackage{subcaption}

\definecolor{drop_green}{RGB}{0,100,0}
\definecolor{bonus_red}{RGB}{180,0,0}
\definecolor{bg}{RGB}{176,226,255}

\newcommand{\todo}[1]{{\textcolor{bonus_red}{#1}}}

\newcommand{\banghua}[1]{{\textcolor{red}{Banghua: #1}}}

\usepackage[capitalize,noabbrev]{cleveref}

\title{Beyond Ten Turns: Unlocking Long-Horizon Agentic Search with Large-Scale Asynchronous RL}

\newcommand{\name}{{ASearcher}}

\author{%
  Jiaxuan Gao$^{1}$,  Wei Fu$^{12}$, Minyang Xie$^{1}$, Shusheng Xu$^{2}$,  \\
  \textbf{Chuyi He$^{2}$, Zhiyu Mei$^{2}$, Banghua Zhu$^{3}$, Yi Wu$^{1}$\thanks{~Corresponding author}} \\
  \\
  $^{1}$ IIIS, Tsinghua University, $^{2}$ Ant Group\\ $^{3}$ University of Washington\\
  \texttt{samjia2000@gmail.com}, \texttt{jxwuyi@gmail.com}\\
}

\begin{document}

\maketitle

\vspace{-3mm}
\begin{abstract}
Recent advancements in LLM-based agents have demonstrated remarkable capabilities in handling complex, knowledge-intensive tasks by integrating external tools. Among diverse choices of tools, search tools play a pivotal role in accessing vast external knowledge. However, open-source agents still fall short of achieving expert-level \emph{Search Intelligence}, the ability to resolve ambiguous queries, generate precise searches, analyze results, and conduct thorough exploration. Existing approaches fall short in scalability, efficiency, and data quality.
For example, small turn limits in existing online RL methods, e.g. $\le 10$, restrict complex strategy learning. 
This paper introduces {\name}, an open-source project for large-scale RL training of search agents. Our key contributions include:
(1) \emph{Scalable fully asynchronous RL training that enables long-horizon search} while maintaining high training efficiency.
(2) \emph{A prompt-based LLM agent that autonomously synthesizes high-quality and challenging QAs.}
Through RL training, our prompt-based QwQ-32B agent achieves substantial improvements. 
Notably, our agent exhibits extreme long-horizon search, with tool calls exceeding 100 turns and output tokens exceeding 400k during training time.  
With a simple agent design and no external LLMs, {\name}-Web-QwQ achieves Avg@4 scores of 51.1 on xBench and 58.7 on GAIA, surpassing existing open-source 32B agents. Finally, we also show that {\name}-Web-QwQ could achieve performance of commercial systems using external summary tool in a zero-shot transfer manner and test-time search. \footnote{We open-source our models, training data, and codes in \href{ASearcher}{https://github.com/inclusionAI/ASearcher}.}

\end{abstract}

\begin{figure}[!h]
    \vspace{-5mm}
    \centering
    \includegraphics[width=0.95\linewidth]{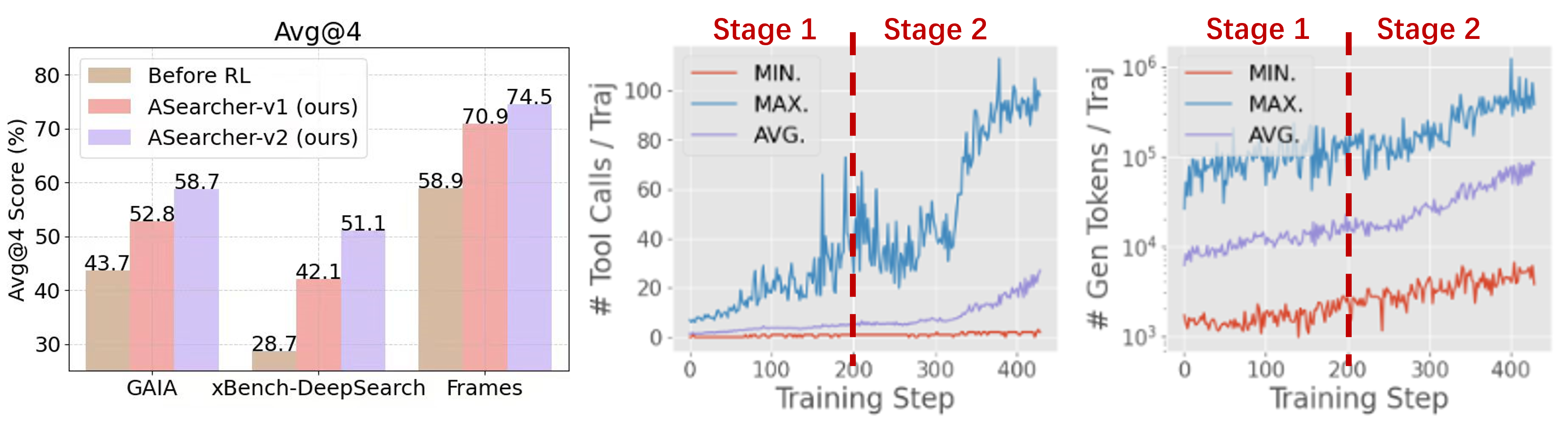}
    \caption{\textbf{(Left) Asynchronous RL brings substantial improvements:} Through RL training, our agent, {\name}-Web-QwQ, obtains +15.0, +22.4, and +15.6 improvements on GAIA, xBench, and Frames, respectively. \textbf{(Middle) \& (Right) Through RL training, {\name}-Web-QwQ learns to conduct long-horizon search}, with tool calls exceeding 100 turns and output tokens exceeding 400k during training. The agent also \textbf{learns expert-level search strategies} (See case study in Sec.~\ref{sec:limit-of-current})}
    \label{fig:main-results}
    \vspace{-2mm}
\end{figure}

\section{Introduction}


{Recent advances in LLM-based agents have demonstrated remarkable capabilities in solving complex, knowledge-intensive problems by leveraging single or multiple external tools~\cite{xi2025rise,yao2023react,wang2024survey}. Among these, \textbf{search tools} stand out as particularly critical, enabling agents to access vast external knowledge for enhanced problem-solving~\cite{openai2025deep,google2025geminideep,perplexity2025deep}. However, expert-level use of search requires advanced intelligence. 
For instance, consider the question \emph{``As of December 31, 2024, what were the numbers of gold, silver, and bronze medals won by China in the 2012 London Olympics?''}.While seemingly straightforward, this query is indeed challenging due to conflicting answers online (e.g., ``38 gold, 27 silver, 22 bronze'' vs. ``39 gold, 31 silver, 22 bronze''). A search agent must navigate noisy and conflicting answers from diverse sources, identify the root cause of conflicts as doping test disqualifications from official reports, and ultimately determine the correct answer. 
Challenging real-world tasks require the agent to resolve high uncertainty in input queries, generate precise search queries, analyze and extract key insights from massive data, resolve inconsistencies, and conduct in-depth exploration. We term this advanced capability \emph{"Search Intelligence".}} 

{Proprietary agents and models has already exhibit signs of complex search behaviors through large-scale Reinforcement Learning (RL) training~\citep{kimi-researcher2025,openai_deep_research}. 
However, open-source approaches for developing search agents still face significant limitations.
A series of works employ \emph{Reinforcement Learning} or \emph{Supervised Fine-Tuning} approaches to incentivize tool-using capabilities~\citep{jin2025search,song2025r1,zheng2025deepresearcher,tan2025rag}. 
On the other hand, \emph{prompt-based LLM agents} supported by open-source models could perform massive tool calls without training~\citep{li2025search,alzubi2025open}.
However, in practice, we find that existing online RL approaches fail to incentivize complex and effective search strategies. We also find prompt-based LLM agents could fail due to the insufficient capabilities of the LLM, such as failing to precisely extract key information from noisy webpages and unable to verify wrong conclusions.
More recently, some works further build up on prompt-based LLM agents, utilizing offline RL approaches to improve the prompt-based agents~\citep{SimpleDeepSearcher,li2025webthinker}. However, this offline RL paradigm, has been shown to underperform online RL in a broader range of domains~\citep{xu2024dposuperiorppollm,fu2021d4rldatasetsdeepdatadriven,song2024trialerrorexplorationbasedtrajectory}.
}

{In reasoning tasks such as math and coding, online RL has enable the models to evolve complex behaviors through iterative refining the reasoning processes based on correctness feedback.~\cite{guo2025deepseek,deepscaler2025,fu2025areal},. This raises a critical question: \emph{How could online RL methods effectively unlock Search Intelligence in open-source agents?}}

{We identify two critical obstacles hindering effective online RL training for search agents:} 
\begin{itemize}

\item \textbf{Insufficient search turns limit complex strategy learning.}  Existing works, such as Search-R1~\cite{jin2025search}, artificially limit the number of search turns, e.g. $\le10$ per trajectory, preventing the agent from exploring deeper search paths. However, complex queries often require multi-turn tool calls and multi-step reasoning, that could not be learned under strict turn limits. 

\item  {\textbf{Lack of large-scale, high-quality question-answer (QA) pairs:} RL training for reasoning tasks requires abundant, challenging, and correct QA pairs~\cite{Polaris2025,li2025questa,yu2025dapo}. However, most existing open-source datasets for search agents are often outdated (e.g. HotpotQA), oversimplified, or too small, failing to stimulate complex search behaviors through RL~\cite{yang2018hotpotqa,li2025websailor,tao2025webshaperagenticallydatasynthesizing}.}
\end{itemize}

\begin{figure}[h]
    \centering
    \includegraphics[width=\linewidth]{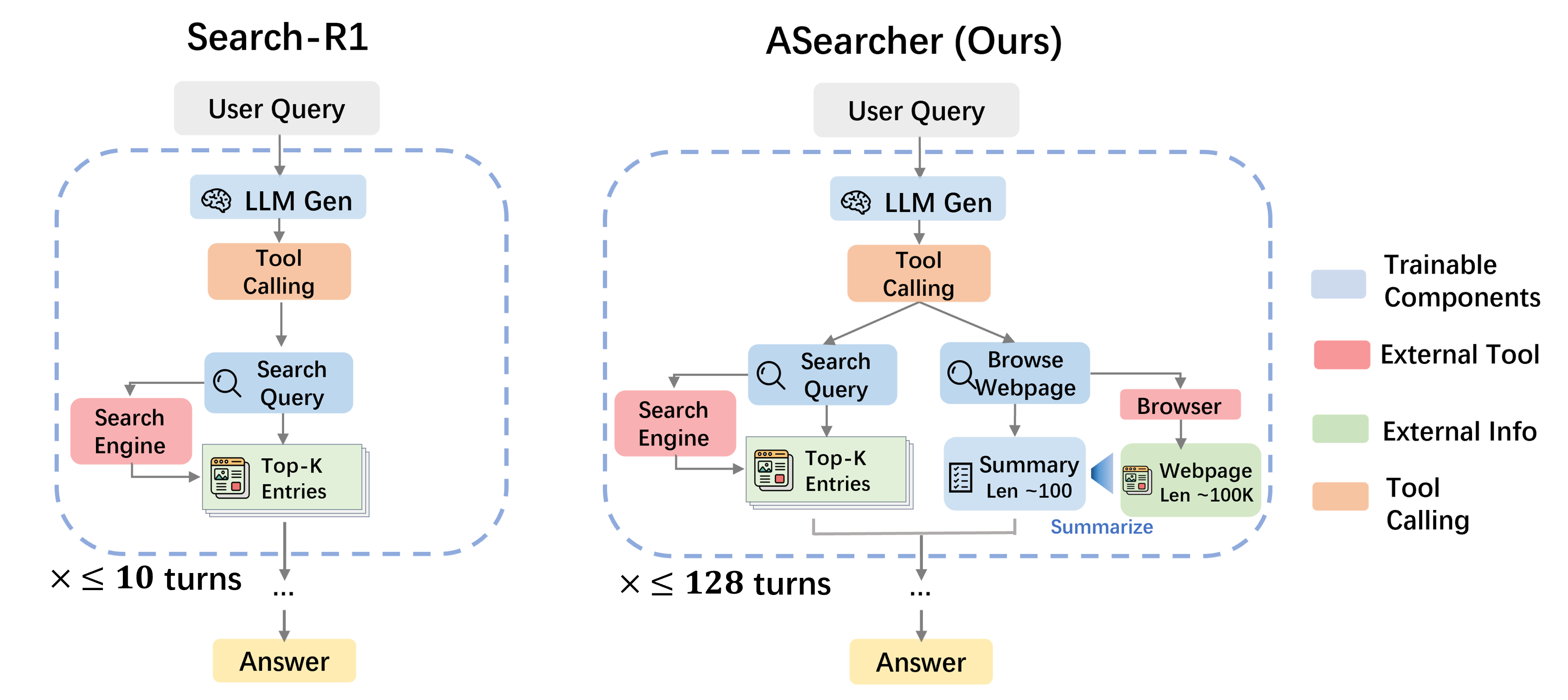}
    \caption{Comparison between {\name} and Search-R1. (Left) Search-R1 is only equipped with search tools and lacks web browsing capability. (Right) {\name} utilizes a simple agent design with two basic tools including search and browsing tools, without relying on any external LLM. {\name} is a comprehensive agent capable of both reasoning and summarizing lengthy web contents. Notably, both reasoning and summarization abilities are optimized through end-to-end RL training.}
    \label{fig:agent_comparison}
\end{figure}

{To address these challenges, we introduce {\name}, an open-source project to enable \emph{large-scale agentic RL training} for search agents. Our contributions include:}

\begin{itemize}
\item {\textbf{Long-horizon search via fully asynchronous agentic RL training.} With a large turn limit in batch generation RL training systems~\citep{jin2025search,song2025r1,deepcoder2025,team2025intellect}, long trajectories within a batch could easily lead to significant idle time, slowing down the whole training process. Building up on AReaL ~\citep{fu2025areal}, our fully asynchronous system avoids long trajectories from blocking the training by decoupling trajectory execution from model updates. This allows relaxed turn limits (e.g., 128 turns/trajectory), enabling agents to explore deeper search paths without sacrificing training efficiency. Remarkably, our agent, {\name}-Web-QwQ, achieves extreme long-horizon search, with tool calls exceeding 100 turns and generated tokens surpassing 400k during RL training. }
\item {\textbf{A scalable QA synthesis agent.} We design an LLM-based agent that autonomously generates \textbf{challenging, uncertain, and grounded} QA pairs requiring multi-turn tool use. Starting from seed questions, the agent iteratively \emph{fuzzes queries} by obscuring key information, or \emph{injects external facts} to increase complexity. Each constructed question undergoes \emph{multi-stage validation} to ensure quality and difficulty. From 14k seed QAs, we generate 134k high-quality samples, with 25.6k requiring external tools for resolution.}
\end{itemize}

Using {\name}, we train agents equipped with search engines and browsers under two settings, \emph{RL training starting from base models} (Qwen2.5-7B/14B), to demonstrate that our training pipeline incentivizes strong and generalizable search strategies, and \emph{fine-tuning a prompt-based agent empowered by a powerful LRM (QwQ-32B)}, to validate the scalability of our training pipeline in fine-tuning large-scale prompt-based LLM agents. 

We evaluate our agents with on multi-hop QA benchmarks and challenging benchmarks including GAIA~\citep{mialon2023gaia}
, xbench-DeepSearch~\citep{xbench}, and Frames~\citep{krishna2024factfetchreasonunified}. 
{\name}-Local-7B/14B, trained only with local knowledge base, demonstrate surprisingly generalizability to realistic web search and achieve state-of-the art performances on multi-hop and single-hop QA tasks. 
Building up on QwQ-32B, {\name}-Web-QwQ achieves an Avg@4 score of 51.1 on xBench-DeepSearch and 58.7 on GAIA, surpassing a set of open-source agents. When evaluating Pass@4, {\name}-Web-QwQ achieves 74.7 on GAIA and 75.0 on xBench-DeepSearch. Notably, through RL training, {\name}-Web-QwQ obtains 78.0\% and 34.3\% improvements on xBench-DeepSearch and GAIA, respectively. 

{\name} presents a large-scale open-source online agentic RL pipeline for LRM-based and LLM-based search agents, unlocking Search Intelligence through scalable training and high-quality data. We hope our findings not only advance search agents but also inspire broader innovations in LLM agents for complex real-world tasks.

\section{Limitations of Existing Open-source Approaches}
\label{sec:limit-of-current}

\begin{figure}
    \centering
\vspace{-2cm}
    \includegraphics[width=0.9\linewidth]{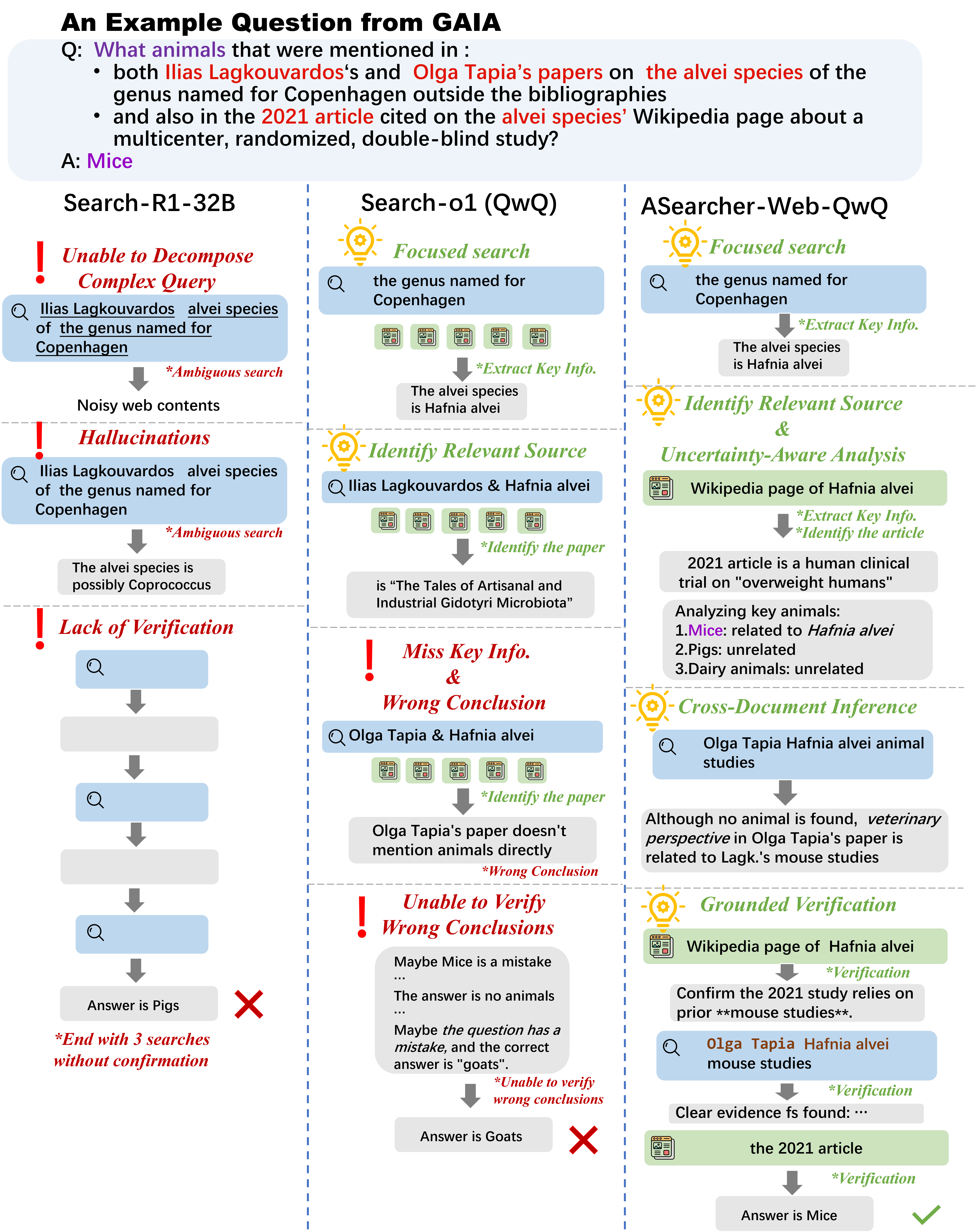}
    \caption{A case study on a complex query from GAIA. \textbf{Search-R1-32B} is unable to break down the complex question and has severe hallucinations. \textbf{Search-o1 (QwQ)} can identify the corrects articles through extensive tool calls, but easily misses key information and fails to verify wrong conclusions. Our end-to-end RL agent, \textbf{{\name}-Web-QwQ}, exhibits key behaviors featuring Search Intelligence: \emph{uncertainty-aware reasoning} (list and examine candidate answers), \emph{precise extraction} from noisy contents, \emph{cross-document inference}, and \emph{grounded verification}.}
    \label{fig:case-study-clean}
\end{figure}

In this section, we provide a detailed case study on an extremely challenging question from GAIA~\citep{mialon2023gaia}. Specifically, we analyze Search-R1-32B~\citep{jin2025search} and Search-o1 (QwQ)~\citep{li2025search} in Fig.~\ref{fig:case-study-clean}. The detailed trajectories are provided in Appendix~\ref{app:case-study}.

\paragraph{Solution Path of the Sample Question.} In Fig.~\ref{fig:case-study-clean}, our case study is carried out on a question requiring finding some specific animal with 4 \textbf{\textcolor{red}{unknown variables}}. To identify the correct answer, the search agent should first find out the mentioned species according to condition ``genus named for Copenhagen'', identify the correct 2021 article based on the citation in the wikipedia page of the species, and then find out the papers of the two mentioned persons. Finally, the correct answer should be determined by cross referencing the 2021 article and the papers. To summarize, this example is challenging for several reasons,
\begin{itemize}
\item \textbf{High Uncertainty:} The question involves multiple unknown variables that could point to many different entities. For example, the 2021 article could point to any article published in 2021 and could only be determined by checking the ``multicenter, randomized, double-blind study'' in the Wikipedia page of the alvei species. 
\item \textbf{Requirement for Exact Information Extraction:} To find the answer, the agent should list all animals mentioned on the webpages and making cross-document comparison. This would require the agent to precisely extract key information from the vast, noisy web contents, instead of simply summarizing the webpages.
\item \textbf{Misleading Answers:} During the process of solving this task, there could be multiple misleading answers, such as "pigs". The agent should rigorously verify its conclusions by checking the intended answer in all related webpages and documents.
\end{itemize}

\paragraph{Existing Online RL Approaches Fail to Learn Complex Search Strategies.} 
In Fig.~\ref{fig:case-study-clean}, Search-R1-32B is not able to decompose the complex query into individual components, consequently only making ambiguous queries that involve too many unknown information. The agent also has severe hallucinations, producing conclusions that are not supported by the search results. Finally, it fails to resolve all unknown information. This case study shows that existing online RL approaches only incentivize elementary search strategies. It is also worth noting that, since the turn limit is set as a small value, e.g. 4, during training, the model only exhibits a short tool-use horizon.

\paragraph{Prompt-based LLM Agents Could Fail Due to Insufficient Capability of the LLM.} In Fig.~\ref{fig:case-study-clean}, Search-o1 (QwQ) can find the species name, as well as the 2021 article and the related papers through a large amount of tool calls. However, when trying to find the answer, Search-o1 (QwQ) would easily miss key information, consequently making incorrect conclusions. Note that even when the agent finds information that directly links to the correct answer, it is still misguided by previous incorrect conclusions. Finally, the agent is unable to verify the correctness of previous conclusions. This case study reveals that, though an open-source model that is not explicitly trained on agentic tasks can perform extensive tool calls, it could not make expert-level reasoning based on the retrieved contents and history contexts.

\paragraph{{\name}-Web-QwQ.} We also analyze the search strategy of our end-to-end RL agent, {\name}-Web-QwQ.
As shown in Fig.~\ref{fig:case-study-clean}, {\name}-Web-QwQ decomposes the complex query into precise queries. Unlike Search-o1 (QwQ) that visits a large amount of websites after each search query, {\name}-Web-QwQ focuses on visiting one website at a time. {\name}-Web-QwQ summarizes all related information from a website. Specifically, all candidate answers are listed and carefully analyzed by the agent. When the search results do not directly point to the desired target, e.g. when searching with ``Olga Tapia Hafnia alvei animal studies'' to find the animals related to Olga Tapia's paper, the agent does not get a clear information but is able to infer the correct answer by making connection with the other paper. After the correct answer ``Mice'' is found, the agent spends further turns on verifying previous conclusions before reporting the final answer. In summary, {\name} successfully train a search agent that exhibits expert-level search behaviors, 
\begin{itemize}
\item \textbf{Uncertainty-aware reasoning:} the agent exhaustively lists and examines all possibilities for uncertain entities
\item \textbf{Precise Key Information Extraction:}  the agent is able to identify the key information from vast, noisy web contents.
\item \textbf{Cross-document Inference:} the agent is able to infer critical conclusions by making connections across multiple documents.
\item \textbf{Grounded Verification:} the agent verifies the correctness of previous conclusions by accessing or searching the related materials. 
\end{itemize}

\section{ASearcher}

In this work, we present {\name}, an open-source project for unlocking search intelligence in search agents through large-scale RL training.
As shown in Fig.~\ref{fig:case-study-clean}, {\name} trains a search agent that is able to solve complex questions by exhaustively resolving all uncertainties and performing multi-turn tool calls.
In the subsequent sections, we present the agent design, the training data as well as data synthesis agent, and fully asynchronous reinforcement learning training in {\name}.

\subsection{Agent Design}

We employ a simple agent design in {\name}, as illustrated in Fig.~\ref{fig:agent_comparison}.
 
\paragraph{Tools.} Given a user query, the agent can utilize two basic tools: \textbf{a search engine} and \textbf{a web browser}.
The search engine takes a text query as input and returns relevant snippets along with their corresponding URLs. In The web browser accepts a URL and returns the content of the webpage.
To effectively tackle complex problems, the model should strategically combine these tools and extract key information from the vast amount of data.

\paragraph{Webpage Summarization.} 
Webpages could contain excessively long contents, therefore we employ the agent to summarize the webpage into a compact summary. 
At training time, this summarization process would also be optimized, allowing the agent to improve the summarization ability through RL training.



\paragraph{Instantiating {\name} with Base LLMs and Advanced LRMs.} Within the framework of {\name}, we investigate two specific instantiations of the search agent: either \emph{base LLMs} such as Qwen2.5-7B/14B, or \emph{advanced Large Reasoning Models (LRMs)} such as QwQ-32B. These two different types of instantiations require different design choices in history management and prompting.

\begin{itemize}
    \item For \textbf{base LLMs}, we following prior works~\citep{jin2025search,song2025r1}, to adopt \emph{append-only} style prompting for the agent. Specifically, starting from a system prompt, all LLM-generated responses, search results and summaries of webpages are appended to the history. The agent takes as input the full history in chronological order and outputs some reasoning texts and actions. This approach ensures efficiency during inference time.
    \item For \textbf{LRMs}, LRMs are already equipped with instruction following capabilities. Therefore we instruct the LRM with different prompts for tool selection, summarization, and answering. We also note that LRMs typically generate long responses, and sometimes the history would be long. We need to ensure a compact input to ensure the LRM generates tokens with a sufficient budget. Therefore, in the history, we discard thinking processes but instead keep summarized thoughts and tool callings. When prompting the LRM, only the most recent 25k characters of the history are provided to the LRM as additional context. These simple designs ensure that the LRM receives an input of at most 10k tokens.
\end{itemize}



\paragraph{End-to-End Reinforcement Learning.} Finally, we highlight that the all LLM-generated responses of the agent, including the thinking process, tool calling, and summarization, are trained using Reinforcement Learning in an end-to-end manner.


\subsection{Training Data}


Our training data are from two primary sources. First, we carefully filter samples from open-source datasets to ensure difficulty and quality. Second, we synthesize high-quality question-answer (QA) pairs specifically designed to guide the agent to learn generalizable search strategies.

\subsubsection{Open-source Data.}
We begin with the training sets from HotpotQA\cite{yang2018hotpotqa} and 2WikiMultiHopQA\cite{ho2020constructingmultihopqadataset}, both of which are multi-hop QA datasets. We employ a model-based filtering process. We first train a model on the full set of open-source data with RL, and then generate 16 responses for each question using the trained model. Finally, we filter out questions that meat any of the following criteria,
\begin{itemize}
\item The model could not find a correct answer out of 16 responses
\item The model achieves $\ge 50\%$ accuracy, meaning the question would not be challenging enough
\item The model finds a correct answer with only a few search turns (i.e., $\le 1$ turns). 
\end{itemize}
This filtering approach ensures we keep only the most challenging yet solvable questions that demand tool use. Finally, from a total of 304k QA pairs, we retain 16k challenging samples for RL training.

Additionally, we include a set of question-answer (QA) pairs designed for accessing certain webpages. In particular, we incorporate a small subset of WebWalkerQA\cite{wu2025webwalker} to help the model learn how to locate answers within noisy, real-world web search environments.

\begin{figure}
\includegraphics[width=\textwidth]{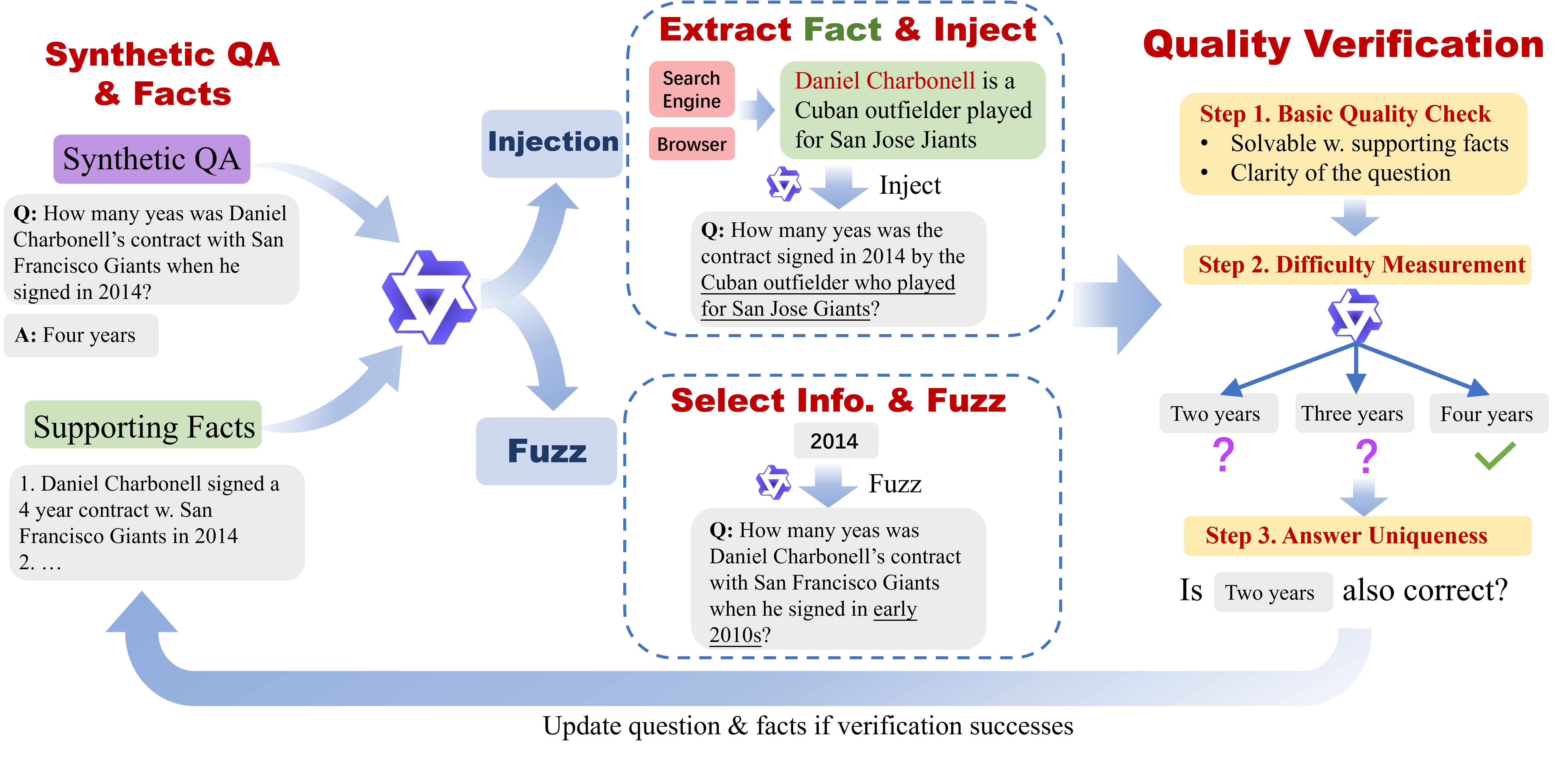}
\caption{Data Synthesis Agent. Starting from a seed QA, the data synthesis agent iteratively modifies the question through two actions, \emph{Injection} and \emph{Fuzz}. Through \emph{injection}, the agent enriches the question by adding some external facts. Through \emph{Fuzz}, the agent blurs certain information to increase uncertainty and difficulty. The related fact to the question are tracked during the synthesis process. Each time the question is modified, a quality verification step is applied to ensure quality and difficulty of the synthetic questions. }
\label{fig:data-synthesis}
\end{figure}

\subsubsection{Data Synthesis Agent}

\begin{figure}[t]
    \centering
    \includegraphics[width=0.9\linewidth]{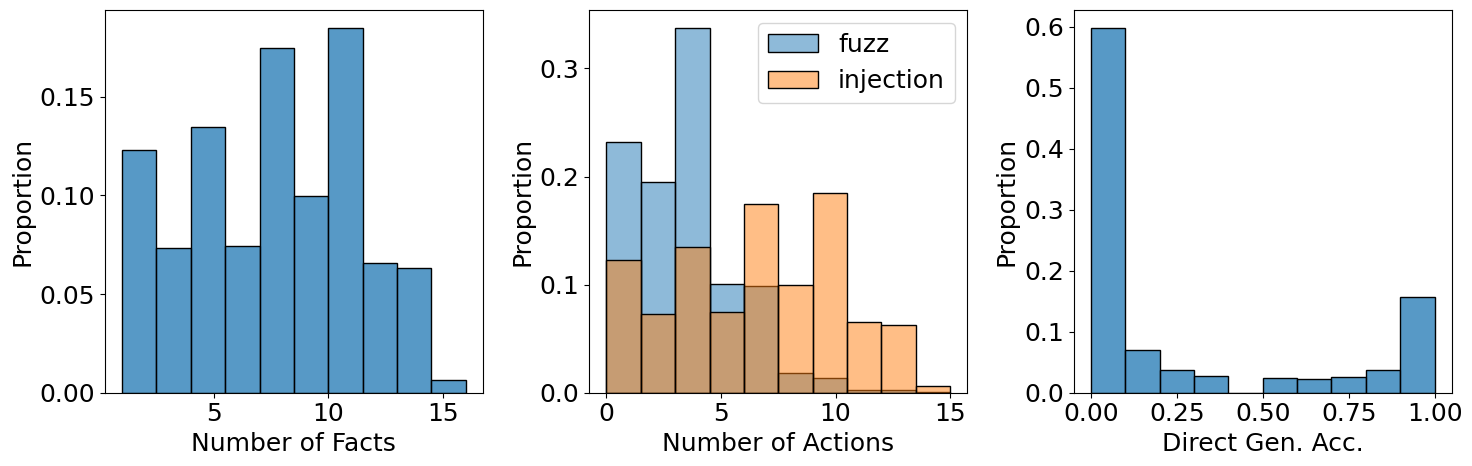}
    \caption{Statistics from our data synthesis process. (Left) The distribution of the number of supporting facts. (Middle) The distribution of the number of fuzz actions and injection actions. (Right) The accuracy distribution of QwQ-32B in answering the generated questions without using any tools.}
    \label{fig:synt_data_stats}
\end{figure}

We further develop a data synthesis agent to create high-quality question-answer pairs. As shown in Fig.~\ref{fig:data-synthesis}, the data synthesis agent begins with a seed question, and iteratively modifies the question to increase the complexity. To ensure the synthetic question is strictly aligned with reliable sources, a list of \emph{supporting facts} obtained during the question synthesis process is kept and continuously updated for quality verification. At each step, given the current question and a list of supporting facts, the agent automatically selects between two key actions,

\begin{table}[h]
\centering
\small
\caption{Examples of the synthetic questions, where \textcolor{red}{red} indicates injected facts and \textcolor{cyan}{cyan} represents fuzzed content.}
\label{tab:question_evolution}
\setlength{\tabcolsep}{3pt}
\begin{tabularx}{\textwidth}{>{\RaggedRight}p{1.5cm}>{\RaggedRight}p{1.5cm}>{\RaggedRight}X}
\toprule
\textbf{Round} & \textbf{Action} & \textbf{Question} \\
\midrule
\textbf{Seed QA} & - & When was Michael P. Hein born? \\
\midrule
\textbf{Round 1} & Injection &  When was the Eckerd College alumnus \textcolor{red}{who served as the first County Executive of Ulster County, New York, and graduated with a Bachelor of Arts in Business Administration} born? \\
\midrule
\textbf{Round 2} & Injection & When was the individual born who, as County Executive of Ulster County, New York, \textcolor{red}{permitted the {Catskill Mountain Railroad} to continue operations between Kingston and Hurley during the 2016 United States House of Representatives elections and also held that position during the 2018 elections}? \\
\midrule
\textbf{Round 3} & Fuzzing & When was the individual born who, as County Executive of Ulster County, New York, permitted \textcolor{cyan}{a historic mountain railway} to continue operations between Kingston and Hurley during the 2016 United States House of Representatives elections and also held that position during the 2018 elections? \\
\midrule
... & ... & ...\\
\midrule
\bottomrule

\toprule
\textbf{Seed QA} & - & Where is the Riggs-Hamilton American Legion Post No. 20 located? \\
\midrule
\textbf{Round 1} & Injection & Where is the American Legion Post \textcolor{red}{in Russellville, Arkansas, built in 1934 and recognized as a notable example of WPA Rustic architecture and listed on the National Register of Historic Places} located? \\
\midrule
\textbf{Round 2} & Fuzzing & Where is the American Legion Post in Russellville, Arkansas, built \textcolor{cyan}{in the early 1930s} and recognized as a notable example of New Deal-era public works architecture and listed on the National Register of Historic Places located? \\
\midrule
\textbf{Round 3} & Fuzzing & Where is the \textcolor{cyan}{veterans' organization's building} in Russellville, Arkansas, built in the early 1930s and recognized as a notable example of New Deal-era public works architecture and listed on the National Register of Historic Places located? \\
\midrule
... & ... & ... \\
\bottomrule
\end{tabularx}

\end{table}

\begin{itemize}
\item \textbf{Action 1: Injection} aims to enrich the context of the question by inserting facts related to the question. The agent first selects an entity in the question and then obtains one piece of related fact about the selected entity from external sources such as Wikipedia. Then a new question is proposed by \emph{injecting} the fact into the question. This injection action increases complexity of the question.


\item \textbf{Action 2: Fuzzing} blurs certain details in the question to increase the uncertainty level of the question. For example, "Catskill Mountain Railroad" could be replaced with "a historic mountain railway". Through fuzzing the question multiple times, both the uncertainty level and difficulty of the question would gradually increase. 
\end{itemize}


To ensure that a synthetic question is of high quality and to precisely evaluate the difficulty, we incorporate a rigorous \textit{quality verification} phase for assessing synthetic questions,

\begin{itemize}
\item \textbf{Step 1. Basic Quality.} We employ an LLM to assess the basic quality of each question. This verification includes checking the clarity of the question and verifying whether the question-answer pair is accurate based on the supporting facts. This quality control step ensures that each question-answer pair is properly grounded in reliable sources.

\item \textbf{Step 2. Difficulty Measurement.} We employ a cutting-edge LRM (e.g., QwQ-32B) to generate multiple answers directly for the synthetic question, without using any external tool. 
This verification process also serves as a measure of question difficulty.

\item \textbf{Step 3. Answer Uniqueness.} The fuzzing action may loosen constraints excessively, compromising the uniqueness of the answer. To prevent ambiguity resulting from multiple correct answers, we evaluate whether any of the mismatched answers generated during the Difficulty Measurement step could serve as alternative valid answers. 
\end{itemize}

We provide two illustrative examples in Tab.~\ref{tab:question_evolution}. Starting with a simple question, the injection action replaces specific entities with related factual details. For instance, ``Michael P. Hein'' is expanded to ``who served as the first County Executive of Ulster County, New York...''. The fuzzing action introduces ambiguity by generalizing precise information, replacing the exact year ``1934'' with ``the early 1930s'' or substituting ``Catskill Mountain Railroad'' with ``a historic mountain railway.''

Through iterative injection and fuzzing, the data synthesis agent produces questions that involve complex information and high uncertainty, requiring extensive search and reasoning to find the correct answer.
After completing the question synthesis process, we filter out questions that the LRM can directly generate the correct answer without relying on search tools. Since these questions can be answered solely based on the intrinsic knowledge of the model, they provide little value for enhancing search capabilities.

Starting with 14,107 seed questions, we perform an average of 6.3 injections and 3.2 fuzzes per question. From the synthetic pool, we select up to three high-quality variations per seed question. This curation process produces a final dataset of 25,624 entries, with the selected questions averaging 4.27 injections and 2.10 fuzzes each.


\subsection{Asynchronous Agentic RL Training}

\subsubsection{Challenges of Scaling Up Trajectory Length in RL} 


In this section, we first empirically show that complex tasks require extensive tool calls and therefore RL training with a large turn limit is necessary for training advanced search agents. Then we show that variance of trajectory execution time is large during training, which could lead to significant idle time in batch generation RL systems.

\begin{figure}[!h]
    \centering
    \begin{minipage}{.33\linewidth}
        \includegraphics[width=\linewidth]{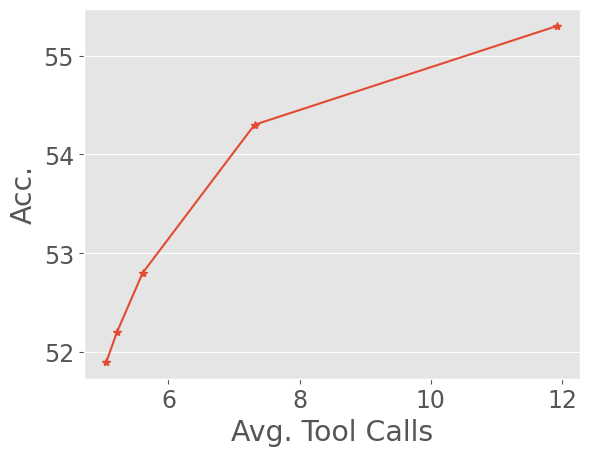}
    \end{minipage}
    \begin{minipage}{.32\linewidth}
        \centering
        \includegraphics[width=\linewidth]{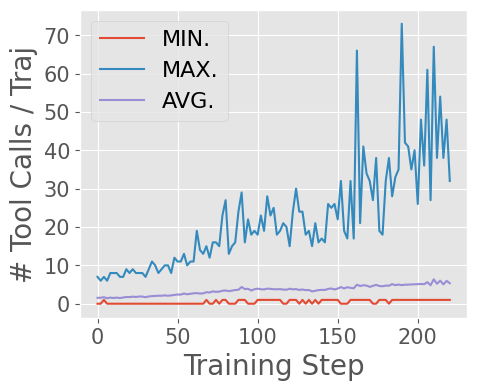}
    \end{minipage}
    \begin{minipage}{.32\linewidth}
        \centering
        \includegraphics[width=\linewidth]{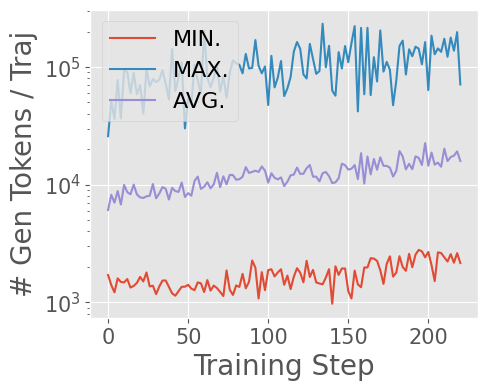}
    \end{minipage}
    \caption{(Left) Test scaling of {\name}-Web-QwQ. 
    Data points are obtained by enforcing different minimum turns.
    The accuracy is averaged over GAIA, xBench-DeepSearch, and Frames. (Middle) Number of tool calls versus training steps. During training time, long trajectories require much more tool calls than short ones. (Right) Number of generated tokens versus training steps. The number of output tokens exhibits significant variance, with long trajectories exceeding short ones by up to two orders of magnitude.}
    \label{fig:acc_wrt_turn}
\end{figure}

\paragraph{Complex Tasks Require Long Trajectories.}
Agentic tasks often require extensive LLM generations and multiple tool calls to solve complex problems, leading to prolonged trajectory execution time.
As shown in Fig.~\ref{fig:acc_wrt_turn}(Left), we evaluate our RL-trained QwQ-32B agent on GAIA~\cite{mialon2023gaia}, xBench-Deepsearch~\cite{krishna2024factfetchreasonunified} and Frames~\citep{krishna2024factfetchreasonunified}, forcing the agent to use tools for different minimal turn numbers. The results demonstrate that accuracy improves with more turns, confirming that complex tasks demand longer trajectories for effective problem-solving.

\paragraph{High Variance in Trajectory Execution Time.}  
Long trajectories also introduce significant variance in execution time.
We analyze the number of tool calls and token generation during RL training of our QwQ agent (Fig.~\ref{fig:acc_wrt_turn}) and observe that the longest trajectories can span dozens more tool calls and two orders of magnitude more tokens than shorter ones. This disparity leads to highly unpredictable per-trajectory runtime, further complicating training efficiency.

\paragraph{Efficiency Issues of Agentic RL Training.}Both prolonged execution and high runtime variance degrade RL training efficiency. We take one-step-off RL training system~\citep{deepcoder2025} as a representative example for batch generation RL systems. In one-step-off RL training, training for step N and trajectory generation for step N+1 are executed concurrently. As shown in Fig.~\ref{fig:time-areal}, though this system overlaps trajectory rollouts with model training, batch generation remains bottlenecked by the slowest trajectory (e.g., trajectory 7), causing GPU idle time and under-utilization.

\subsubsection{Fully Asynchronous RL Training.}
To ensure efficient agentic RL training, we adopt a fully asynchronous training paradigm. Notably, our approach incorporates asynchornization at the two distinct aspects. 

\paragraph{Asynchronous Trajectory Rollouts.} Trajectory rollouts are collected in parallel and do not directly interfere with each other. Each trajectory independently sends tool calling requests to corresponding servers and LLM generation requests to the LLM inference engine. Concurrent requests from different trajectories are automatically handled by the servers. Fully independent trajectory execution ensures a trajectory does not need to wait for other trajectories when generating LLM responses and waiting for tool calling responses, thereby improving training efficiency.

\begin{figure}
\centering
\includegraphics[width=1.0\textwidth]{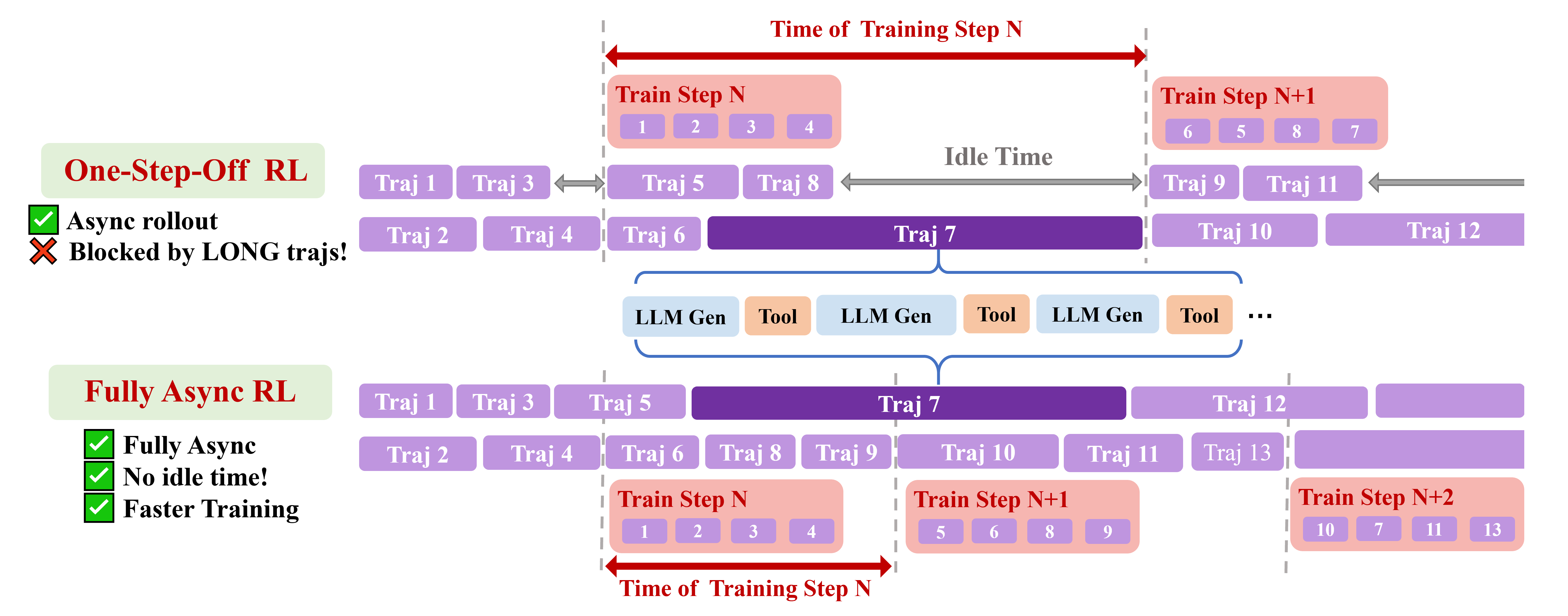}
\caption{One-Step-off RL v.s. Fully Asynchronous RL. In batch generation systems, a batch should wait for the longest trajectory, leading to significant GPU idle time. In contrast, fully asynchronous RL achieves faster training than batch generation RL by fully decoupling training and trajectory generation, achieving near-full resource utilization for trajectory generation.}
\label{fig:time-areal}
\end{figure}

\paragraph{Decoupled Rollout and Training.} Besides asynchronous rollout, trajectory rollouts and model updates are also fully decoupled. In Fig.~\ref{fig:time-areal}, we compare our fully asynchronous RL training with one-step-off RL training, which utilizes asynchronous rollout within batches. 
In fully asynchronous RL training, long trajectories do not block generation and can span multiple versions, significantly reducing GPU idle time and achieving near-full GPU utilization during generation. On the training side, a training step is launched as soon as sufficient trajectories are collected to form a batch. As shown in Fig.~\ref{fig:time-areal}, the training process does not wait for the extremely long trajectory 7 but instead proceeds with trajectory 9.

\subsection{Training Details}

\paragraph{MDP Formulation.} We follow the formulation of Markov Decision Process (MDP). Formally, an MDP is defined by the tuple $(S, A, T, R)$. Here $S$ represents the state space, usually containing the history, search results, and retrieved webpages. $A$ denotes the action space and an action includes tokens generated by the agent. Some tool calling could be extracted from the action through specific tags, e.g. <search> search query </search>. $T(s'|s,a)$ is the transition probability, where $s'$ is the updated state after applying the tool calling in action $a$ at state $s$.
At each timestep, the agent receives a state $s_t$ and generates an action $a_t$ with policy $\pi:S\rightarrow A$. The goal of the agent is to maximize the return $J(\pi) = \mathbb{E}\left[\sum_{t=0}^{\infty} R(s_t, a_t) \bigg| a_t \sim \pi(s_t)\right]$.

\paragraph{GRPO Training.} We employ the GRPO~\citep{shao2024deepseekmath} algorithm to train search agents. Specifically, for each input question $x$, $G$ trajectories $\tau_1,\tau_2,\cdots,\tau_G$ are generated where $\tau_i=(s_0^i,a^i_0,s_1^i,\cdots,s_{T_i}^i)$. To optimize the agent, we employ the following loss,
\begin{align}
\mathcal J_{GRPO}(\theta)=\mathbb E_{x\sim \mathcal D,\{\tau_i\}_{i=1}^G\sim\pi_{\theta_{old}}(\cdot|x)}\Bigg[&\frac{1}{G}\sum_{i=1}^G\frac{1}{\sum_{t=0}^{T_i-1}|a^i_t|}\sum_{t=0}^{T_i-1}\sum_{j=1}^{|a_t^i|}\min\Bigg( \frac{\pi_\theta(a_{t,j}^i|s_t,a_{t,<j}^i)}{\pi_{\theta_{old}}(a_{t,j}^i|s_t,a_{t,<j}^i)}\hat A_{i},\nonumber \\
&\text{clip}\Bigg(\frac{\pi_\theta(a_{t,j}^i|s_t,a_{t,<j}^i)}{\pi_{\theta_{old}}(a_{t,j}^i|s_t,a_{t,<j}^i)},1-\epsilon,1+\epsilon\Bigg)\hat A_{i}\Bigg)
\Bigg]
\end{align}

where $\epsilon$ is a hyperparameter,  and $\hat A_{i}$ is the advantage for the $i$-th trajectory, computed based on the relative rewards of all trajectories within each group.

\paragraph{Dynamic Filtering.} To enhance training efficiency, we implement dynamic filtering to exclude queries that lack meaningful training signals. Specifically, we remove queries where all responses yield identical rewards (resulting in zero advantages), including both queries where the agent already achieves high accuracy and those with incorrectly labeled answers.

\paragraph{Reward Function.} For reward function, we adopt a sparse-reward setting where rewards are computed at trajectory completion. When training from base LLMs, the reward function combines a format reward and F1 score through multiplication. When fine-tuning LRM-based agents (e.g., QwQ), we utilize LLM-as-Judge\cite{liu2023calibrating}\cite{wang2024leave} as the reward function and omit format rewards, as these models inherently maintain proper output formatting.

\section{Experiments}
\subsection{Experiment Setup}

\paragraph{Benchmarks.} We first evaluate the agents on single-hop and multi-hop QA tasks. For single-hop questions, we use  Natural Questions~\citep{naturalquestions}, TriviaQA~\citep{joshi2017triviaqa} and PopQA~\citep{mallen2023llm_memorization}. For multi-hop questions, we use HotpotQA~\citep{yang2018hotpotqa}, 2WikiMultiHopQA~\citep{ho2020constructingmultihopqadataset}, MuSiQue~\citep{trivedi2022musique}, and  Bamboogle~\citep{press2022measuring}. We further perform evaluation on more challenging benchmarks including Frames~\citep{krishna2024factfetchreasonunified}, GAIA~\citep{mialon2023gaia}, and xBench-DeepSearch~\citep{xbench} as extra test sets. We evaluate our approach on 1000 randomly sampled instances from the validation sets of HotpotQA, 2WikiMultiHopQA, and MuSiQue. For Bamboogle, Frames, GAIA and xBench-DeepSearch, we use their full test sets. For GAIA, we use the 103 examples from the text-only validation subset~\citep{li2025search}.

\paragraph{Search Tools.} We evaluate the search agents with two settings, each with different types of search tools. In the first setting, \textbf{local knowledge base with RAG}, agents interact with a locally deployed RAG system to retrieve related information from a Wikipedia 2018 corpus~\citep{karpukhin2020dense}. 
In the other \textbf{web-based search and browsing} setting, agents operate in an interactive web environment with access to both a search engine and a browser tool. 
For more challenging benchmarks, GAIA, xBench-DeepSearch and Frames, we only conduct evaluations under this web-based setting.

\paragraph{Baselines} We consider two groups of baselines aligned with the two benchmark categories. For the multi-hop and single-hop QA benchmarks, we include Search-R1(7B/14B/32B)~\citep{jin2025search}, R1-Searcher(7B)~\citep{song2025r1}, Search-o1(QwQ-32B)~\citep{li2025search}, DeepResearcher~\citep{zheng2025deepresearcher} and SimpleDeepSearcher~\citep{SimpleDeepSearcher}. We also prompt Qwen-2.5-7B/32B to directly generate answers without using any tools. On the more challenging benchmarks, we compare against powerful 32B-scale models, including direct generation with QwQ-32B, Search-o1(QwQ-32B)~\citep{li2025search}, Search-R1-32B~\citep{jin2025search}, WebThinker-QwQ~\citep{li2025webthinker},SimpleDeepSearcher-QwQ~\citep{SimpleDeepSearcher} and WebDancer-32B~\citep{wu2025webdancer}. All baselines are evaluated using the same tools as our agent to ensure a fair comparison. 

\paragraph{Evaluation Metrics} We adopt two complementary evaluation metrics: F1 score and LLM-as-Judge (LasJ). The F1 score is computed at the word level, measuring the harmonic mean of precision and recall between the predicted and reference answers. 
For LLM-as-Judge, a strong LLM (Qwen2.5-72B-Instruct) is prompted to assess the correctness of model outputs according to task-specific instructions. On GAIA, xBench-DeepSearch and Frames, we only use LLM-as-Judge and report the Avg@4 and Pass@4 scores for all models. 

\paragraph{Training Details of {\name}.} We set the turn limit as 32 for 7B and 14B models, and 128 for {\name}-Web-QwQ. The batch size is set as 128 for 7B and 14B models, and 64 for {\name}-Web-QwQ. We curate two sets of training data, one for 7B/14B training and the other for QwQ-32B training. These two datasets are both of 35k sizes and open-sourced. Training of {\name}-Web-QwQ takes approximated 16k H800 GPU hours in total.

\paragraph{Two-Stage Curriculum.} For {\name}-Web-QwQ, we employ a two-stage curriculum RL training process. In the first stage, the training data covers questions with varying difficulty levels, including simple questions that are solvable with a small number of tool calls. In the second stage, we retain only the questions that require at least 5 tool calls to solve to further activate the long-horizon agentic capability of the model. The final checkpoint of these two stages are name {\name}-Web-QwQ${\text{-v1}}$ and {\name}-Web-QwQ${\text{-v2}}$, respectively.

\subsection{Main Results}

We present the main experiment results across three evaluation settings: (1) local knowledge base with retrieval-augmented generation (RAG) on standard QA benchmarks, (2) web-based search and browsing on the same benchmarks, and (3) web-based search and browsing on more challenging benchmarks. \textbf{{\name}}, instantiated with \text{Qwen2.5-7B}, \text{Qwen2.5-14B}, and \text{QwQ-32B}, consistently outperforms existing opensource agents of the same model scale on both F1 and LasJ metrics. {\name}-14B achieves the best performance across 7B, 14B, and 32B models on a suite of multi-hop and single-hop QA benchmarks, and {\name}-QwQ significantly outperforms several strong baselines of comparable size on these challenging benchmarks. These results highlight the generality and scalability of {\name} across diverse tasks and model sizes.

\begin{table}[]
    \centering
    \scriptsize
    \setlength{\tabcolsep}{3pt}
    \caption{Results with Local Knowledge Base.}
    \label{tab:results_local}
    \begin{tabular}{l|cccccccc|cccccc|cc}
    \toprule
        \multirow{3}{*}{Method}
        & \multicolumn{8}{c|}{Multi-Hop QA} & \multicolumn{6}{c|}{Single-Hop QA} & \multicolumn{2}{c}{Avg.} \\
    \cmidrule{2-15}
        & \multicolumn{2}{c}{2WikiMQA} & \multicolumn{2}{c}{HotpotQA} & \multicolumn{2}{c}{Bamboogle} & \multicolumn{2}{c|}{Musique} & \multicolumn{2}{c}{NQ} & \multicolumn{2}{c}{TriviaQA} & \multicolumn{2}{c|}{PopQA}  & \\
        & F1 & LasJ & F1 & LasJ & F1 & LasJ & F1 & LasJ & F1 & LasJ & F1 & LasJ & F1 & LasJ &  F1 & LasJ \\
    \midrule
        \multicolumn{17}{c}{\textbf{7B Models}} \\
    \midrule
        Qwen-2.5-7B Direct Gen. & 30.4 & 29.4 & 29.2 & 30.9 & 37.2 & 42.4 & 11.8 & 11.0 & 27.9 & 29.4 & 50.4 & 59.8 & 21.5 & 20.5 & 29.8 & 31.9\\
        Search-R1-7B & 54.7 & 58.1 & 57.6 & 60.8 & \textbf{55.8} & 58.4 & 28.2 & 27.1 & \textbf{58.7} & 49.9 & 68.0 & 78.0 & 57.3 & 55.7 & 54.3 & 55.4 \\
        R1-Searcher-7B & 64.0 & 67.1 & 57.1 & 61.0 & 51.8 & 56.0 & 28.7 & 27.3 & 51.2 & 49.1 & 62.0 & 72.8 & 50.9 & 49.5 & 52.2 & 54.7 \\
    \arrayrulecolor{gray!50}\midrule\arrayrulecolor{black}
    \rowcolor{green!15}
        {\name}-Local-7B & \textbf{72.3} & \textbf{77.6} & \textbf{62.6} & \textbf{67.6} & 55.0 & \textbf{60.0} & \textbf{34.4}& \textbf{32.6} & 55.6 & \textbf{54.5} & \textbf{68.1} & \textbf{79.3} & \textbf{57.9} & \textbf{55.9} & \textbf{58.0} & \textbf{61.0} \\
        
    \midrule
        \multicolumn{17}{c}{\textbf{14B/32B Models}} \\
    \midrule
        QwQ-32B Direct Gen. & 34.6 & 35.4 & 37.1 & 40.2 & 56.9 & 61.6 & 16.8 & 16.1 & 36.9 & 38.2 & 65.4 & 75.8 & 27.9 & 26.3 & 39.4 & 41.9 \\
        Search-R1-14B & 48.2 & 49.8 & 56.2 & 58.9 & 52.8 & 51.2 & 27.0 & 25.7 & 60.0 & 51.2 & 71.0 & 79.9 & 56.1 & 54.3 & 53.0 & 53.0 \\
        Search-R1-32B & 63.1 & 67.5 & 60.5 & 64.0 & \textbf{60.0} & 61.6 & 34.4 & 32.9 & \textbf{60.8} & 52.2 & \textbf{72.0} & 82.1 & \textbf{60.3} & \textbf{58.2} & 58.7 & 59.8 \\
    \arrayrulecolor{gray!50}\midrule\arrayrulecolor{black}
    \rowcolor{green!15}
        {\name}-Local-14B & \textbf{72.2} & \textbf{79.1} & \textbf{65.1} & \textbf{71.0} & 59.4 & \textbf{64.8} & \textbf{35.6} & \textbf{34.6} & 56.6 & \textbf{56.1} & 71.6 & \textbf{84.0} & 57.6 & 55.9 & \textbf{59.7} & \textbf{63.6}  \\
    \bottomrule
    \end{tabular}

\end{table}

\paragraph{Local Knowledge Base with RAG on Standard QA Benchmarks.}
As shown in Table~\ref{tab:results_local}, {\name}-Local, trained via reinforcement learning with local knowledge base, achieves the best performance across 7B and 14B on a suite of multi-hop and single-hop QA benchmarks.
In the 7B setting, {\name} attains an average F1 of \textbf{58.0}, outperforming strong baselines such as Search-R1-7B (54.3) and R1-Searcher-7B (52.2). It also achieves a LasJ score of \textbf{61.0}, significantly outperforming Search-R1-7B (55.4) and R1-Searcher-7B (54.7).
The gains are even more pronounced at the 14B scale, where {\name}-Local-14B reaches an F1 of \textbf{60.0} and LasJ of \textbf{65.6}, surpassing even the larger 32B retrieval-based baseline Search-R1-32B.

\begin{table}[]
    \centering
    \scriptsize
    \setlength{\tabcolsep}{2pt}
    \caption{Results with Web-based Search and Browsing.}
    \label{tab:results_web}
    \begin{tabular}{lc|cccccccc|cccccc|cc}
    
    \toprule
        \multirow{3}{*}{Method} & \multirow{3}{*}{\makecell{Training\\Setting}} 
        & \multicolumn{8}{c|}{Multi-Hop QA} & \multicolumn{6}{c|}{Single-Hop QA} & \multicolumn{2}{c}{Avg.} \\
    \cmidrule{3-16}
        & & \multicolumn{2}{c}{2WikiMQA} & \multicolumn{2}{c}{HotpotQA} & \multicolumn{2}{c}{Bamboogle} & \multicolumn{2}{c|}{Musique} & \multicolumn{2}{c}{NQ} & \multicolumn{2}{c}{TriviaQA} & \multicolumn{2}{c|}{PopQA}  & \\
        & & F1 & LasJ & F1 & LasJ & F1 & LasJ & F1 & LasJ & F1 & LasJ & F1 & LasJ & F1 & LasJ &  F1 & LasJ \\
    \midrule
        \multicolumn{18}{c}{\textbf{7B Models}} \\
    \midrule
        Qwen-2.5-7B Direct Gen. & - & 30.8 & 30.9 & 28.6 & 29.5 & 37.2 & 39.6 & 10.6 &  1.9 & 29.6 & 29.9 & 51.2 & 59.3 & 19.8 & 17.4 & 29.7 & 29.8 \\ 
        Search-R1-7B & local & 58.9 & 64.8 & 59.0 & 62.8 & 66.3 & 73.6 & 29.4 & 25.4 & \textbf{58.4} & 51.1 & 73.1 & 84.1 & \textbf{53.0} & \textbf{51.3} & 56.9 & 59.0 \\
        R1-Searcher-7B & local & 66.6 & 69.4 & 56.8 & 61.6 & 62.8 & 72.0 & 28.7 & 25.3 & 49.6 & 48.7 & 67.6 & 79.5 & 46.5 & 45.2 & 54.1 & 57.4 \\
        DeepResearcher-7B & web & 61.0 & 64.1 & 57.1 & 61.0 & \textbf{68.8} & \textbf{76.8} & 26.8 & 24.5 & 52.0 & 52.9 & 70.0 & 82.8 & 48.9 & 45.7 & 54.9 & 58.3  \\
        Simple DS-7B & web & 67.4 & 73.9 & 57.6 & 62.5 & 61.5 & 72.0 & 26.4 & 26.2 & 43.9 & 53.1 & 73.9 & 85.4 & 43.7 & 48.8 & 53.5 &  60.3 \\
    \arrayrulecolor{gray!50}\midrule\arrayrulecolor{black}
    \rowcolor{green!15}
        {\name}-Local-7B & local & \textbf{69.1} & \textbf{75.5} & 61.6 & 67.1 & 66.2 & 76.0 & \textbf{33.3} & \textbf{30.7} & 54.7 & \textbf{53.7} & \textbf{75.2} & \textbf{87.3} & 52.9 & 49.7 & \textbf{59.0} & \textbf{62.9} \\
    \rowcolor{green!15}
        {\name}-Web-7B & web & 67.5 & 73.3 & \textbf{61.7} & \textbf{67.2} & 66.4 & 72.0 & 32.9 & 29.6 & 55.2 & 55.4 & 74 & 85.7 & 52.4 & 48.9 & 58.6 & 61.7  \\
    \midrule
        \multicolumn{18}{c}{\textbf{14B/32B Models}} \\
    \midrule
        QwQ-32B Direct Gen. &  - & 33.7 & 33.4 & 39.1 & 42.1 & 56.9 & 57.9 & 18.8 & 19.3 & 37.8 & 43.0 & 63.8 & 74.2 & 25.9 & 24.5 & 39.4 & 42.1 \\
        Search-o1 (QwQ-32B) & - & 68.9 & 77.8 & 58.4 & 65.3 & 68.6 & 82.4 & 31.8 & 33.5 & 43.1 & \textbf{57.2}& \textbf{76.3} & \textbf{89.6} & 43.2 & 48.3 & 55.8 & 64.9  \\
        Search-R1-14B & local & 51.8 & 53.8 & 55.3 & 58.6 & 67.4 & 75.2 & 29.8 & 26.9 & 57.7 & 49.6 & 74.4 & 83.9 & 51.0 & 49.8 & 55.4 & 56.8 \\
        Search-R1-32B & local & 63.7 & 69.3 & 60.3 & 64.2 & \textbf{76.4} & \textbf{81.6} & 33.0 & 30.8 & \textbf{58.6} & 51.1 & 76.2 & 86.6 & \textbf{55.0} & \textbf{53.6} & 60.4 & 62.5 \\
        Simple DS-QwQ & web & 71.7 & 80.4 & 62.0 & 67.5 & 73.2 & 83.2 & 33.3 & 32.9 & 45.7 & 55.3 & 77.2 & 90.2 & 45.5 & 47.8 & 58.4 & 65.3 \\
        \rowcolor{red!15}
    \arrayrulecolor{gray!50}\midrule\arrayrulecolor{black}
    \rowcolor{green!15}
        {\name}-Local-14B & local & 70.4 & 79.8 & \textbf{63.6} & 70.5 & 68.7 & 80.8 & 35.1 & 33.8 & 53.5 & 55.4 & 76.1 & 88.5 & 52.5 & 50.5 & 60.0 & \textbf{65.6} \\
    \rowcolor{green!15}
        {\name}-Web-14B & web & \textbf{76.1} & \textbf{80.7} & 63.5 & \textbf{68.5} & 69.9 & 75.2 & \textbf{36.6} & \textbf{33.7} & 56.0 & 55.5 & 75.4 & 87.6 & 52.9 & 50.0 & \textbf{61.5} & 64.5 \\   
    \bottomrule
    \end{tabular}
\end{table}

\paragraph{Web-based Search and Browsing on Standard QA Benchmarks} 
In Table~\ref{tab:results_web}, we evaluate agents in a realistic web-based setting.
Notably, we evaluate models trained entirely with local knowledge base in the web setting in a zero-shot manner, to directly examine the generalizability of search strategies learned through RL. 
Across both model sizes, {\name} consistently outperforms strong baselines. In particular, {\name}-Web-14B achieves the best performance with an average F1 of \textbf{61.5}, surpassing SimpleDeepSearcher, the strongest 32B baseline in this setting. Remarkably, {\name}-Local-14B model exhibits strong generalization when tested in the web-based setting, achieving significant gains over all baseline models of similar or larger size in terms of LasJ. This confirms that {\name} learns generalizable search strategies that transfer to different sources of information.

\begin{table}[]
    \centering
    \small
    \caption{Results on GAIA, xBench-DeepSearch, and Frames. The results are evaluated with LLM-as-Judge. For baselines, we run the corresponding official codes for 4 seeds and report Avg@4 and Pass@4.}
    \label{tab:main}
    \begin{tabular}{l|cc|cc|cc}
    \toprule
       \multirow{2}{*}{Method} & \multicolumn{2}{c|}{GAIA} & \multicolumn{2}{c|}{xBench-DeepSearch} & \multicolumn{2}{c}{Frames} \\
        & Avg@4 & Pass@4 & Avg@4 & Pass@4 & Avg@4 & Pass@4 \\
    \midrule 
            QwQ-32B Direct Gen.  & 23.1 & 31.1 & 11.8 & 23.0 & 29.9 & 39.9 \\ 
            Search-o1 (QwQ)  & 48.1 & 67.0 & 40.3 & 65.0 & 63.6 & 81.1 \\
            Search-R1-32B  & 28.6 & 43.7 & 19.5 & 37.0 & 44.1 & 61.0 \\ 
            WebThinker-QwQ  & 42.5 & 57.3 & 32.8 & 52.0 & 57.7 & 79.5 \\ 
            Simple DS-QwQ  & 47.6 & 64.1 & 35.8 & 61.0 & 67.0 & 82.2 \\
            WebDancer-QwQ  & 47.4 & 61.2 & 40.0 & \textbf{68.0} & 63.8 & 81.4 \\ 
    \arrayrulecolor{gray!50}\midrule\arrayrulecolor{black}
     \rowcolor{green!15}
            {\name}-Web-QwQ${\text{-v1}}$ & \textbf{52.8} & \textbf{70.1} & \textbf{42.1} & \textbf{68.0} & \textbf{70.9}  & \textbf{84.0} \\
     \rowcolor{green!15}
            {\name}-Web-QwQ${\text{-v2}}$ & \textbf{58.7} & \textbf{74.7} & \textbf{51.1} & \textbf{75.0} & \textbf{74.5}  & \textbf{85.5} \\
            
    \bottomrule
    \end{tabular}
    
\end{table}

\paragraph{Web-based Search and Browsing on Challenging Benchmarks.}
Table~\ref{tab:main} shows experiment results on challenging QA tasks that require advanced problem-solving capabilities and search strategies. These benchmarks are specifically designed to assess the agent’s ability to interact with real web and retrieve up-to-date information that often go beyond the internal knowledge of LLMs. As a result, direct generating answers from models (e.g., QwQ-32B) perform poorly across all datasets.
Our agent, {\name}-Web-QwQ, achieves the best Avg@4 scores on GAIA (58.7) and xBench-DeepSearch (51.1), outperforming previous state-of-the-art open-source agents. These results further highlight superiority in handling long-horizon planning, real-world tool use, and open-domain exploration. Besides Avg@4, we also report the Pass@4 score that computes the ratio of questions that an agent finds the correct answer out of 4 trials. {\name}-Web-QwQ also outperforms state-of-the-art open-source agents in terms of pass rate.

\begin{figure}[t]
    \centering
    \includegraphics[width=0.8\linewidth]{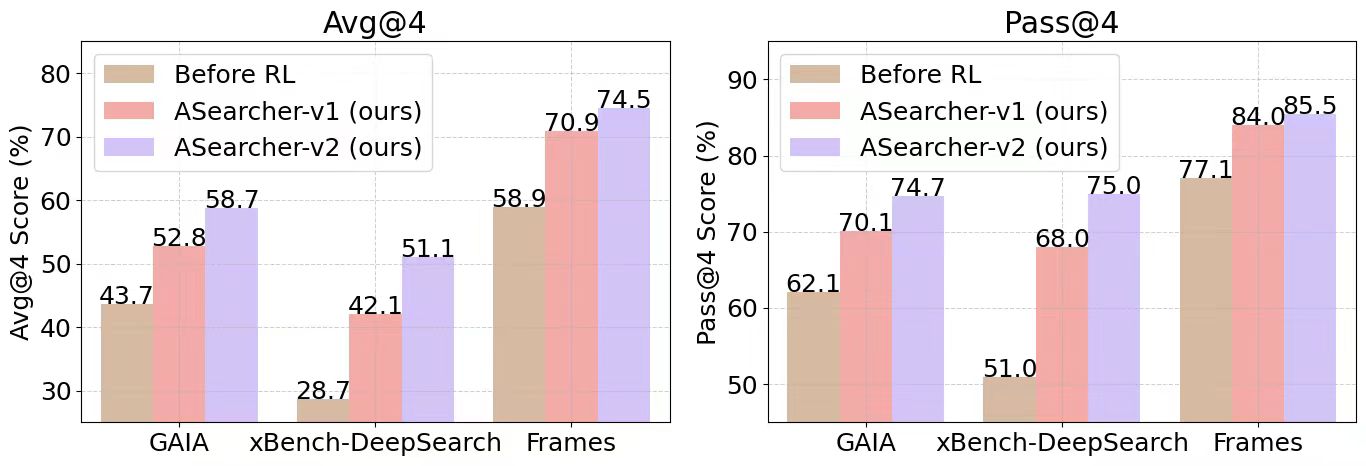}
    \caption{Comparison of the performance of QwQ-32B agent before and after RL Training.}
    \label{fig:before_after_rl}
\end{figure}

\paragraph{Effect of RL Training.} As shown in Fig.~\ref{fig:before_after_rl}, {\name}-Web-QwQ obtains +15.0, +22.4, and +14.6 improvements on GAIA, xBench-DeepSearch and Frames respectively. When considering the pass rate, i.e. Pass@4, {\name}-Web-QwQ also obtains significant gains, especially on xBench-DeepSearch with 24.0 improvements. Significant improvements in pass rate demonstrate that our training pipeline trains the agent to learn complex search strategies to perform precise searches, extract key information, and resolve conflict information. 

\begin{table}[h]
    \centering
    \small
    \caption{Pass@1 results of {\name}-Web-QwQ-v2 and baselines, evaluated on GAIA~\citep{mialon2023gaia}, xBench-DeepSearch~\citep{xbench}, Frames~\citep{krishna2024factfetchreasonunified}, and HLE-500~\citep{li2025webthinker}. $^\dagger$ indicates results are obtained from official reports. } 
    \label{tab:main}
    \resizebox{0.8\textwidth}{!}{
    \begin{tabular}{l|cccc}
    \toprule
       Method  & GAIA & xBench-DeepSearch & Frames & HLE-500\\
    \midrule 
        \multicolumn{5}{c}{\textbf{Commercial Deep Research Agents}} \\
    \midrule
        Kimi-Researcher & - & 69.0$^{\dagger}$ & 78.8$^{\dagger}$ & 26.9$^{\dagger}$ \\
        OpenAI DeepResearch & 67.0$^{\dagger}$ & 26.6$^{\dagger}$ \\
    \midrule 
        \multicolumn{5}{c}{\textbf{General LLMs using Tools}} \\
    \midrule 
        OpenAI-o3 & 70.5$^{\dagger}$ & 66.7 $^{\dagger}$& 84.0$^{\dagger}$ & 20.2$^{\dagger}$ \\
        Qwen3-30B-A3B & 35.9$^{\dagger}$ & 32.0$^{\dagger}$ & 56.4$^{\dagger}$ & 13.2$^{\dagger}$ \\
        Qwen3-235B-A22B & 45.6$^{\dagger}$ & 46.0$^{\dagger}$ & - & 20.0$^{\dagger}$ \\
        DeepSeek-R1 & - & 55.0$^{\dagger}$ & 82.0$^{\dagger}$ & 24.8$^{\dagger}$ \\
        Claude-4-Sonnet & 68.3$^{\dagger}$ & 64.6$^{\dagger}$ & 80.7$^{\dagger}$ & 20.3$^{\dagger}$ \\
        \midrule
        \multicolumn{5}{c}{\textbf{{\name}-Web-QwQ (ours)}} \\
    \midrule
     \rowcolor{green!15}
            {\name}-Web-QwQ-v2 & 58.7 & 51.1 & 74.5  & 21.5 \\
     \rowcolor{green!15}
            + Summary=DeepSeek-V3  & 60.3 & 56.4 & 76.6 & 23.4 \\
     \rowcolor{green!15}
            + Test-time Search (K=16)  & 71.8 & 75.0 & 83.4 & 24.6 \\
    \bottomrule
    \end{tabular}}
    
\end{table}

\paragraph{Zero-Shot Transfer with Summary Tool \& Test-time Scaling.} Although we adopt a single-model configuration during training time, the agent design of {\name} is generalizable. Specifically, the webpage summarization process could be supported by external models. We here use DeepSeek-V3 as the webpage summarization model. From Table~\ref{tab:main}, it is clear that using a more powerful model for summarization improves the performance of {\name}. We further investigate test-time scaling. Specifically, we run $K=16$ independent runs for each problem in the test set and aggregate the conclusions from these independent runs with DeepSeek-V3. As shown in Table~\ref{tab:main}, test-time scaling approach leads to competitive performance with commercial systems, including Kimi-Researcher, OpenAI DeepResearch, and OpenAI o3.

\subsection{Training Dynamics}

\begin{figure}[h]
    \centering
    \begin{subfigure}[t]{0.32\textwidth}
        \centering
        \includegraphics[width=\linewidth]{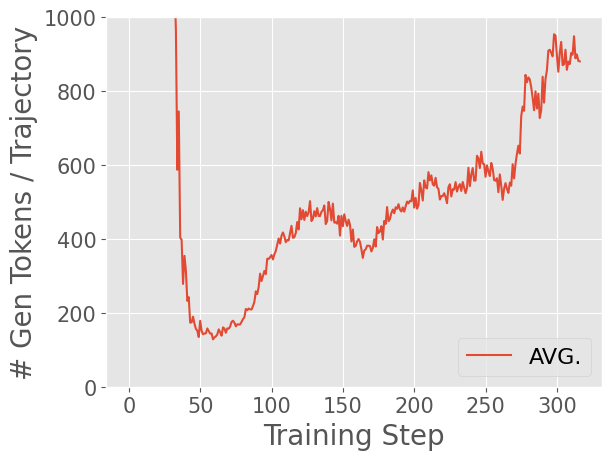}
        \caption{Generated Tokens}
        \label{fig:2a}
    \end{subfigure}
    \hfill
    \begin{subfigure}[t]{0.3\textwidth}
        \centering
        \includegraphics[width=\linewidth]{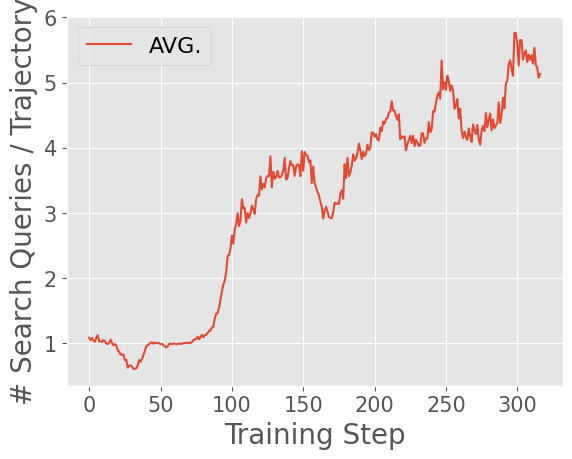}
        \caption{Search Queries}
        \label{fig:2b}
    \end{subfigure}
    \hfill
    \begin{subfigure}[t]{0.32\textwidth}
        \centering
        \includegraphics[width=\linewidth]{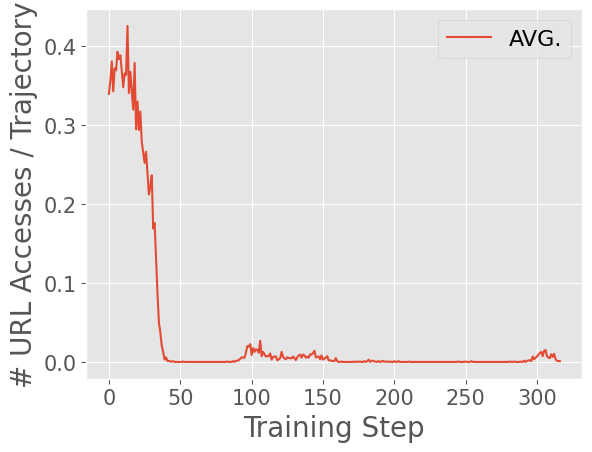}
        \caption{URL Accesses}
        \label{fig:2c}
    \end{subfigure}
    \caption{Training Dynamics of {\name}-Local-7B.}
    \label{fig:local-7b-training}
\end{figure}

\begin{figure}[b]
    \centering
    \begin{subfigure}[t]{0.32\textwidth}
        \centering
        \includegraphics[width=\linewidth]{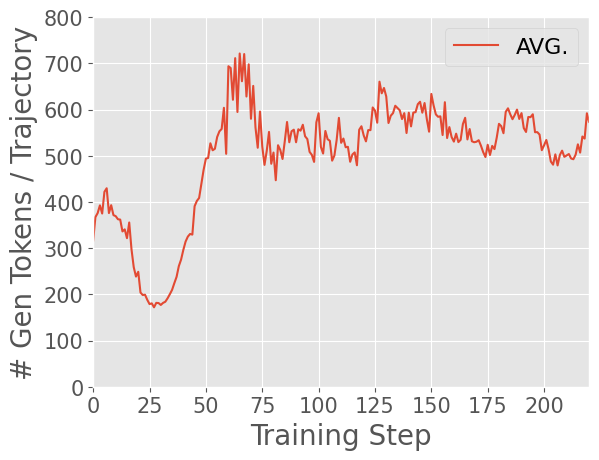}
        \caption{Generated Tokens}
        \label{fig:2a}
    \end{subfigure}
    \hfill
    \begin{subfigure}[t]{0.3\textwidth}
        \centering
        \includegraphics[width=\linewidth]{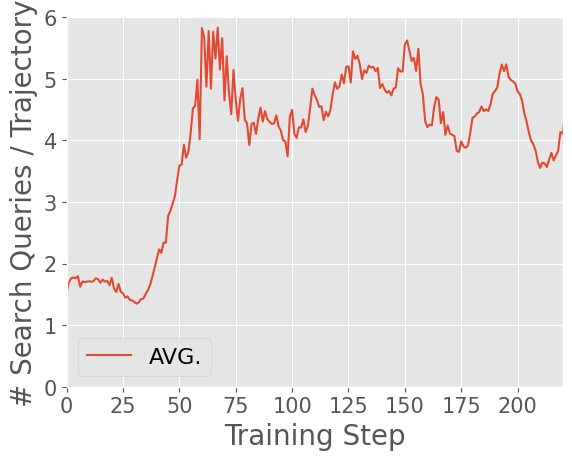}
        \caption{Search Queries}
        \label{fig:2b}
    \end{subfigure}
    \hfill
    \begin{subfigure}[t]{0.32\textwidth}
        \centering
        \includegraphics[width=\linewidth]{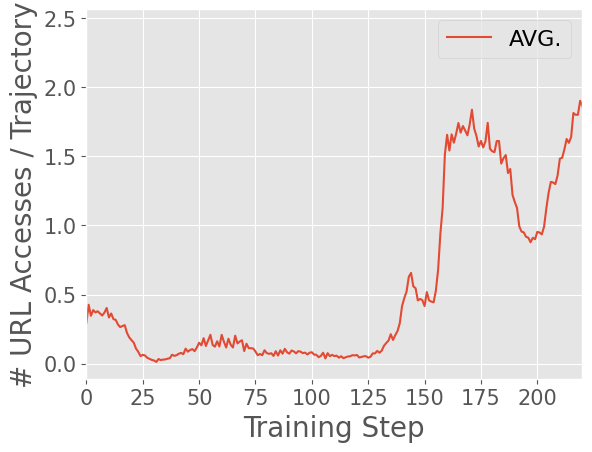}
        \caption{URL Accesses}
        \label{fig:2c}
    \end{subfigure}
    \caption{Training Dynamics of {\name}-Local-14B.}
    \label{fig:local-14b-training}
\end{figure}

\paragraph{Training Dynamics of {\name}-Local-7B/14B.} In Fig.~\ref{fig:local-7b-training} and Fig.~\ref{fig:local-14b-training}, we plot the number of generated tokens, search queries and webpage browsing for {\name}-Local-7B and {\name}-Local-14B training, respectively. With our training recipe, length increment and the increase in the number of tool callings are observed in both 7B and 14B scales. Notably, the number of search queries scale up to 6, which is higher than the numbers reported by prior works~\citep{jin2025search,song2025r1}. Interestingly, we find that the 7B model fails to learn valid webpage browsing while the 14B model can learn to access webpage to solve challenging questions in the later stage of training. We hypothesize that the failure of 7B model in learning webpage browsing occurs because the model capacity is too small to stably learning summarize lengthy webpages in a zero RL training setting.

\paragraph{Training Dynamics of {\name}-Web-QwQ.} Similarly, the training dynamics of {\name}-Web-QwQ are illustrated in Fig.~\ref{fig:acc_wrt_turn}. As the training progresses, the agent learns to perform more tool calls, reaching a maximum of around 40 calls at the 200th step, with peak instances even achieving up to 70 calls. Also the QwQ-32B agent generates more tokens through training, with a maximum of over 150k tokens. This scaling trend in both tool utilization and output length highlights the potential of fully asynchronous RL training for complex real-world agent applications.

\subsection{Emergent Behaviors}

\begin{figure}[h]
    \centering
    \includegraphics[width=0.42\linewidth]{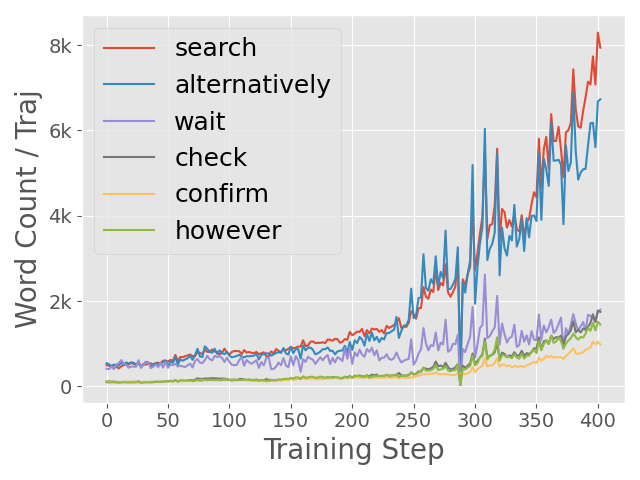}
    \includegraphics[width=0.42\linewidth]{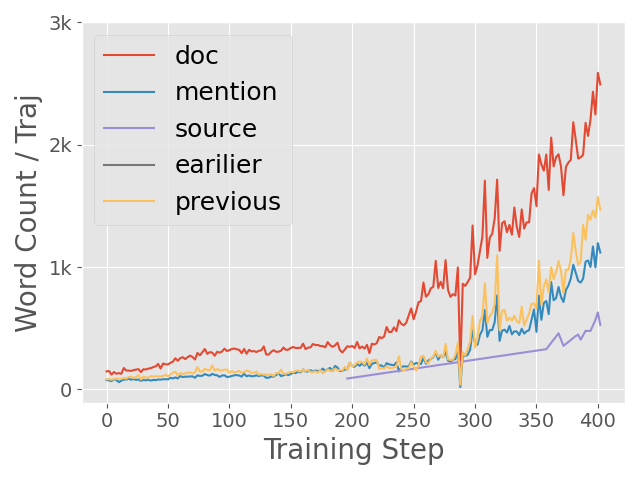}
    \caption{Left: Word count of reflective keywords during training time. Right: Word count of keywords indicating explicit reference of external information.}
    \label{fig:word-count-training}
    \vspace{-5mm}
\end{figure}

\paragraph{Keyword Analysis.} In Fig.~\ref{fig:word-count-training}, we plot the word count of different keywords during the training process of {\name}-Web-QwQ. These keywords include words that indicate reflection behaviors of the agent, such as ``alternatively'', ``however'', and ``wait'', and also words indicating that the agent is referencing to external information, such as ``doc'' and ``mention''. From Fig.~\ref{fig:word-count-training}, it is clear that the agent learns to reflect over previous actions and conclusions. Interestingly, we also see that, in the second stage of RL training where the training focuses on challenging tasks, the agent learns to refer to external information, as evidenced by the increment in the word count of ``doc'' and ``mention'' after step 200.

\section{Related Works}


\paragraph{Search Agents.} 
Some works have constructed agent workflows that enable large language models (LLMs) to leverage external tools for solving complex tasks, with notable examples include Search-o1\cite{li2025search} and ReAgent\cite{zhao2025reagent}.
Prompt-based methods, while effective for rapid development, are fundamentally limited by the capacity of the underlying LLMs and could not be reliably improved with environment feedback. Some works attempt to construct SFT trajectories for LLMs. For instance, \cite{asai2023self,yu2024auto} leverage large LLMs to synthesize retrieval and reasoning trajectories to fine-tune smaller models. 
Recently, some works investigate Reinforcement learning (RL) methods to enhance the LLM-based agents, mostly focusing on multi-hop QA benchmarks such as HotpotQA and 2WikiMultihop. \cite{jin2025search, song2025r1, chen2025learning, zheng2025deepresearcher} perform RL training with multi-hop QA data and observe an increase in the number of tool uses. RAG-R1~\citep{tan2025rag} further combines SFT and RL to enhance the search strategies.  
More recently, researchers have begun to focus on more challenging tasks, by fine-tuning sophisticated prompt-based agents powered by Large Reasoning Models through offline RL~\citep{li2025webthinker}, SFT on simulated trajectories with real-world web data~\citep{SimpleDeepSearcher, li2025websailor}, and constructing challenging QAs for RL training.~\citep{tao2025webshaperagenticallydatasynthesizing}.


\paragraph{Synthetic Data for Search Agents.} 
Rather than relying solely on large-scale human annotation, data synthesis has emerged as a scalable approach to prepare training data for search agents. Recent approaches generate synthetic but realistic QA trajectories by interacting with real web pages and curating data using LRMs~\cite{SimpleDeepSearcher, wu2025webdancer, li2025websailor}. On the other hand, WebSailor~\citep{li2025websailor} constructs structurally challenging tasks through sampling and fuzzing, and WebShaper~\citep{tao2025webshaperagenticallydatasynthesizing} utilizes techniques from set theory to construct high-quality complex QAs. By contrast, {\name} develops an autonomous LLM agent for synthesizing challenging QAs with high uncertainty, without relying on complex knowledge graphs. Both the data synthesis agent and the synthetic training data in {\name} are fully open-sourced.



\section{Conclusion}

In this work, we present {\name}, a open-source project for large-scale RL training. Our contribution includes a fully asynchronous agentic RL training system and a data synthesis agent for large-scale high-quality QA construction. By instantiating {\name} with base LLMs including Qwen2.5-7B/14B and prompt-based LLM agents based on QWQ-32B, {\name} outperforms the state-of-the-art open-source agents across different model sizes and evaluation settings. With fully asynchronous agentic RL training and insight from our data synthesis pipeline, we hope our work could benefit future work on training advanced agents for a broader range of applications.

{
\bibliographystyle{plainnat}
\bibliography{reference}

\begin{thebibliography}{49}
\providecommand{\natexlab}[1]{#1}
\providecommand{\url}[1]{\texttt{#1}}
\expandafter\ifx\csname urlstyle\endcsname\relax
  \providecommand{\doi}[1]{doi: #1}\else
  \providecommand{\doi}{doi: \begingroup \urlstyle{rm}\Url}\fi

\bibitem[AI(2025)]{kimi-researcher2025}
Moonshot AI.
\newblock Kimi-researcher: End-to-end rl training for emerging agentic capabilities.
\newblock \url{https://moonshotai.github.io/Kimi-Researcher/}, 2025.
\newblock URL \url{https://moonshotai.github.io/Kimi-Researcher/}.

\bibitem[Alzubi et~al.(2025)Alzubi, Brooks, Chiniya, Contente, von Gerlach, Irwin, Jiang, Kaz, Nguyen, Oh, et~al.]{alzubi2025open}
Salaheddin Alzubi, Creston Brooks, Purva Chiniya, Edoardo Contente, Chiara von Gerlach, Lucas Irwin, Yihan Jiang, Arda Kaz, Windsor Nguyen, Sewoong Oh, et~al.
\newblock Open deep search: Democratizing search with open-source reasoning agents.
\newblock \emph{arXiv preprint arXiv:2503.20201}, 2025.

\bibitem[An et~al.()An, Xie, Li, Li, Zhang, Gong, Zhong, Xu, Qiu, Wang, and Kong]{Polaris2025}
Chenxin An, Zhihui Xie, Xiaonan Li, Lei Li, Jun Zhang, Shansan Gong, Ming Zhong, Jingjing Xu, Xipeng Qiu, Mingxuan Wang, and Lingpeng Kong.
\newblock Polaris: A post-training recipe for scaling reinforcement learning on advanced reasoning models.
\newblock URL \url{https://hkunlp.github.io/blog/2025/Polaris}.

\bibitem[Asai et~al.(2023)Asai, Wu, Wang, Sil, and Hajishirzi]{asai2023self}
Akari Asai, Zeqiu Wu, Yizhong Wang, Avirup Sil, and Hannaneh Hajishirzi.
\newblock Self-rag: Self-reflective retrieval augmented generation.
\newblock In \emph{NeurIPS 2023 workshop on instruction tuning and instruction following}, 2023.

\bibitem[Chen et~al.(2025)Chen, Li, Sun, Zhou, Zhu, Wang, Pan, Zhang, Chen, Yang, et~al.]{chen2025learning}
Mingyang Chen, Tianpeng Li, Haoze Sun, Yijie Zhou, Chenzheng Zhu, Haofen Wang, Jeff~Z Pan, Wen Zhang, Huajun Chen, Fan Yang, et~al.
\newblock Learning to reason with search for llms via reinforcement learning.
\newblock \emph{arXiv preprint arXiv:2503.19470}, 2025.

\bibitem[Fu et~al.(2021)Fu, Kumar, Nachum, Tucker, and Levine]{fu2021d4rldatasetsdeepdatadriven}
Justin Fu, Aviral Kumar, Ofir Nachum, George Tucker, and Sergey Levine.
\newblock D4rl: Datasets for deep data-driven reinforcement learning, 2021.
\newblock URL \url{https://arxiv.org/abs/2004.07219}.

\bibitem[Fu et~al.(2025)Fu, Gao, Shen, Zhu, Mei, He, Xu, Wei, Mei, Wang, Yang, Yuan, and Wu]{fu2025areal}
Wei Fu, Jiaxuan Gao, Xujie Shen, Chen Zhu, Zhiyu Mei, Chuyi He, Shusheng Xu, Guo Wei, Jun Mei, Jiashu Wang, Tongkai Yang, Binhang Yuan, and Yi~Wu.
\newblock Areal: A large-scale asynchronous reinforcement learning system for language reasoning, 2025.
\newblock URL \url{https://arxiv.org/abs/2505.24298}.

\bibitem[{Google Team}(2025)]{google2025geminideep}
{Google Team}.
\newblock Introducing {Gemini} deep research, 2025.
\newblock URL \url{https://gemini.google/overview/deep-research/}.
\newblock Accessed: 2025-04-06.

\bibitem[Guo et~al.(2025)Guo, Yang, Zhang, Song, Zhang, Xu, Zhu, Ma, Wang, Bi, et~al.]{guo2025deepseek}
Daya Guo, Dejian Yang, Haowei Zhang, Junxiao Song, Ruoyu Zhang, Runxin Xu, Qihao Zhu, Shirong Ma, Peiyi Wang, Xiao Bi, et~al.
\newblock Deepseek-r1: Incentivizing reasoning capability in llms via reinforcement learning.
\newblock \emph{arXiv preprint arXiv:2501.12948}, 2025.

\bibitem[Ho et~al.(2020)Ho, Nguyen, Sugawara, and Aizawa]{ho2020constructingmultihopqadataset}
Xanh Ho, Anh-Khoa~Duong Nguyen, Saku Sugawara, and Akiko Aizawa.
\newblock Constructing a multi-hop qa dataset for comprehensive evaluation of reasoning steps, 2020.
\newblock URL \url{https://arxiv.org/abs/2011.01060}.

\bibitem[Jin et~al.(2025)Jin, Zeng, Yue, Yoon, Arik, Wang, Zamani, and Han]{jin2025search}
Bowen Jin, Hansi Zeng, Zhenrui Yue, Jinsung Yoon, Sercan Arik, Dong Wang, Hamed Zamani, and Jiawei Han.
\newblock Search-r1: Training llms to reason and leverage search engines with reinforcement learning.
\newblock \emph{arXiv preprint arXiv:2503.09516}, 2025.

\bibitem[Joshi et~al.(2017)Joshi, Choi, Weld, and Zettlemoyer]{joshi2017triviaqa}
Mandar Joshi, Eunsol Choi, Daniel~S Weld, and Luke Zettlemoyer.
\newblock Triviaqa: A large scale distantly supervised challenge dataset for reading comprehension.
\newblock \emph{arXiv preprint arXiv:1705.03551}, 2017.

\bibitem[Karpukhin et~al.(2020)Karpukhin, Oguz, Min, Lewis, Wu, Edunov, Chen, and Yih]{karpukhin2020dense}
Vladimir Karpukhin, Barlas Oguz, Sewon Min, Patrick~SH Lewis, Ledell Wu, Sergey Edunov, Danqi Chen, and Wen-tau Yih.
\newblock Dense passage retrieval for open-domain question answering.
\newblock In \emph{EMNLP (1)}, pages 6769--6781, 2020.

\bibitem[Krishna et~al.(2024)Krishna, Krishna, Mohananey, Schwarcz, Stambler, Upadhyay, and Faruqui]{krishna2024factfetchreasonunified}
Satyapriya Krishna, Kalpesh Krishna, Anhad Mohananey, Steven Schwarcz, Adam Stambler, Shyam Upadhyay, and Manaal Faruqui.
\newblock Fact, fetch, and reason: A unified evaluation of retrieval-augmented generation, 2024.
\newblock URL \url{https://arxiv.org/abs/2409.12941}.

\bibitem[Kwiatkowski et~al.(2019)Kwiatkowski, Palomaki, Redfield, Collins, Parikh, Alberti, Epstein, Polosukhin, Devlin, Lee, Toutanova, Jones, Kelcey, Chang, Dai, Uszkoreit, Le, and Petrov]{naturalquestions}
Tom Kwiatkowski, Jennimaria Palomaki, Olivia Redfield, Michael Collins, Ankur~P. Parikh, Chris Alberti, Danielle Epstein, Illia Polosukhin, Jacob Devlin, Kenton Lee, Kristina Toutanova, Llion Jones, Matthew Kelcey, Ming{-}Wei Chang, Andrew~M. Dai, Jakob Uszkoreit, Quoc Le, and Slav Petrov.
\newblock Natural questions: a benchmark for question answering research.
\newblock \emph{Trans. Assoc. Comput. Linguistics}, 7:\penalty0 452--466, 2019.
\newblock \doi{10.1162/tacl\_a\_00276}.
\newblock URL \url{https://doi.org/10.1162/tacl\_a\_00276}.

\bibitem[Li et~al.(2025{\natexlab{a}})Li, Lu, Wen, Yang, Gao, Lin, Wu, and Zhang]{li2025questa}
Jiazheng Li, Hong Lu, Kaiyue Wen, Zaiwen Yang, Jiaxuan Gao, Hongzhou Lin, Yi~Wu, and Jingzhao Zhang.
\newblock Questa: Expanding reasoning capacity in llms via question augmentation.
\newblock \emph{arXiv preprint arXiv:2507.13266}, 2025{\natexlab{a}}.

\bibitem[Li et~al.(2025{\natexlab{b}})Li, Zhang, Yin, Zhang, Ou, Wu, Yin, Li, Tao, Wang, et~al.]{li2025websailor}
Kuan Li, Zhongwang Zhang, Huifeng Yin, Liwen Zhang, Litu Ou, Jialong Wu, Wenbiao Yin, Baixuan Li, Zhengwei Tao, Xinyu Wang, et~al.
\newblock Websailor: Navigating super-human reasoning for web agent.
\newblock \emph{arXiv preprint arXiv:2507.02592}, 2025{\natexlab{b}}.

\bibitem[Li et~al.(2025{\natexlab{c}})Li, Dong, Jin, Zhang, Zhou, Zhu, Zhang, and Dou]{li2025search}
Xiaoxi Li, Guanting Dong, Jiajie Jin, Yuyao Zhang, Yujia Zhou, Yutao Zhu, Peitian Zhang, and Zhicheng Dou.
\newblock Search-o1: Agentic search-enhanced large reasoning models.
\newblock \emph{arXiv preprint arXiv:2501.05366}, 2025{\natexlab{c}}.

\bibitem[Li et~al.(2025{\natexlab{d}})Li, Jin, Dong, Qian, Zhu, Wu, Wen, and Dou]{li2025webthinker}
Xiaoxi Li, Jiajie Jin, Guanting Dong, Hongjin Qian, Yutao Zhu, Yongkang Wu, Ji-Rong Wen, and Zhicheng Dou.
\newblock Webthinker: Empowering large reasoning models with deep research capability.
\newblock \emph{arXiv preprint arXiv:2504.21776}, 2025{\natexlab{d}}.

\bibitem[Liu et~al.(2023)Liu, Yang, Huang, Zhang, Huang, Wei, Deng, Sun, and Zhang]{liu2023calibrating}
Yuxuan Liu, Tianchi Yang, Shaohan Huang, Zihan Zhang, Haizhen Huang, Furu Wei, Weiwei Deng, Feng Sun, and Qi~Zhang.
\newblock Calibrating llm-based evaluator.
\newblock \emph{arXiv preprint arXiv:2309.13308}, 2023.

\bibitem[Luo et~al.(2025{\natexlab{a}})Luo, Tan, Huang, Patel, Ariyak, Wu, Shi, Xin, Cai, Weber, Zhang, Li, Popa, and Stoica]{deepcoder2025}
Michael Luo, Sijun Tan, Roy Huang, Ameen Patel, Alpay Ariyak, Qingyang Wu, Xiaoxiang Shi, Rachel Xin, Colin Cai, Maurice Weber, Ce~Zhang, Li~Erran Li, Raluca~Ada Popa, and Ion Stoica.
\newblock Deepcoder: A fully open-source 14b coder at o3-mini level.
\newblock \url{https://pretty-radio-b75.notion.site/DeepCoder-A-Fully-Open-Source-14B-Coder-at-O3-mini-Level-1cf81902c14680b3bee5eb349a512a51}, 2025{\natexlab{a}}.
\newblock Notion Blog.

\bibitem[Luo et~al.(2025{\natexlab{b}})Luo, Tan, Wong, Shi, Tang, Roongta, Cai, Luo, Li, Popa, and Stoica]{deepscaler2025}
Michael Luo, Sijun Tan, Justin Wong, Xiaoxiang Shi, William~Y. Tang, Manan Roongta, Colin Cai, Jeffrey Luo, Li~Erran Li, Raluca~Ada Popa, and Ion Stoica.
\newblock Deepscaler: Surpassing o1-preview with a 1.5b model by scaling rl.
\newblock \url{https://pretty-radio-b75.notion.site/DeepScaleR-Surpassing-O1-Preview-with-a-1-5B-Model-by-Scaling-RL-19681902c1468005bed8ca303013a4e2}, 2025{\natexlab{b}}.
\newblock Notion Blog.

\bibitem[Mallen et~al.(2022)Mallen, Asai, Zhong, Das, Hajishirzi, and Khashabi]{mallen2023llm_memorization}
Alex Mallen, Akari Asai, Victor Zhong, Rajarshi Das, Hannaneh Hajishirzi, and Daniel Khashabi.
\newblock When not to trust language models: Investigating effectiveness and limitations of parametric and non-parametric memories.
\newblock \emph{arXiv preprint}, 2022.

\bibitem[Mialon et~al.(2023)Mialon, Fourrier, Wolf, LeCun, and Scialom]{mialon2023gaia}
Gr{\'e}goire Mialon, Cl{\'e}mentine Fourrier, Thomas Wolf, Yann LeCun, and Thomas Scialom.
\newblock Gaia: a benchmark for general ai assistants.
\newblock In \emph{The Twelfth International Conference on Learning Representations}, 2023.

\bibitem[{OpenAI}()]{openai_deep_research}
{OpenAI}.
\newblock Openai deep research.
\newblock \url{https://openai.com/index/introducing-deep-research/}.

\bibitem[{OpenAI}(2025)]{openai2025deep}
{OpenAI}.
\newblock Introducing deep research, 2025.
\newblock URL \url{https://openai.com/index/introducing-deep-research/}.
\newblock Accessed: 2025-04-06.

\bibitem[{Perplexity Team}(2025)]{perplexity2025deep}
{Perplexity Team}.
\newblock Introducing {Perplexity} deep research, 2025.
\newblock URL \url{https://www.perplexity.ai/hub/blog/introducing-perplexity-deep-research}.
\newblock Accessed: 2025-04-06.

\bibitem[Press et~al.(2022)Press, Zhang, Min, Schmidt, Smith, and Lewis]{press2022measuring}
Ofir Press, Muru Zhang, Sewon Min, Ludwig Schmidt, Noah~A Smith, and Mike Lewis.
\newblock Measuring and narrowing the compositionality gap in language models.
\newblock \emph{arXiv preprint arXiv:2210.03350}, 2022.

\bibitem[Shao et~al.(2024)Shao, Wang, Zhu, Xu, Song, Bi, Zhang, Zhang, Li, Wu, et~al.]{shao2024deepseekmath}
Zhihong Shao, Peiyi Wang, Qihao Zhu, Runxin Xu, Junxiao Song, Xiao Bi, Haowei Zhang, Mingchuan Zhang, YK~Li, Yang Wu, et~al.
\newblock Deepseekmath: Pushing the limits of mathematical reasoning in open language models.
\newblock \emph{arXiv preprint arXiv:2402.03300}, 2024.

\bibitem[Song et~al.(2025)Song, Jiang, Min, Chen, Chen, Zhao, Fang, and Wen]{song2025r1}
Huatong Song, Jinhao Jiang, Yingqian Min, Jie Chen, Zhipeng Chen, Wayne~Xin Zhao, Lei Fang, and Ji-Rong Wen.
\newblock R1-searcher: Incentivizing the search capability in llms via reinforcement learning.
\newblock \emph{arXiv preprint arXiv:2503.05592}, 2025.

\bibitem[Song et~al.(2024)Song, Yin, Yue, Huang, Li, and Lin]{song2024trialerrorexplorationbasedtrajectory}
Yifan Song, Da~Yin, Xiang Yue, Jie Huang, Sujian Li, and Bill~Yuchen Lin.
\newblock Trial and error: Exploration-based trajectory optimization for llm agents, 2024.
\newblock URL \url{https://arxiv.org/abs/2403.02502}.

\bibitem[Sun* et~al.(2025)Sun*, Song*, Wang, Ren, Jiang, Zhang, Fang, Wang, and Wayne Xin~Zhao]{SimpleDeepSearcher}
Shuang Sun*, Huatong Song*, Yuhao Wang, Ruiyang Ren, Jinhao Jiang, Junjie Zhang, Lei Fang, Zhongyuan Wang, and Ji-Rong~Wen Wayne Xin~Zhao.
\newblock Simpledeepsearcher: Deep information seeking via web-powered reasoning trajectory synthesis.
\newblock 2025.
\newblock URL \url{https://github.com/RUCAIBox/SimpleDeepSearcher}.

\bibitem[Tan et~al.(2025)Tan, Huang, Wu, Zhang, Zhuang, and Gu]{tan2025rag}
Zhiwen Tan, Jiaming Huang, Qintong Wu, Hongxuan Zhang, Chenyi Zhuang, and Jinjie Gu.
\newblock Rag-r1: Incentivize the search and reasoning capabilities of llms through multi-query parallelism.
\newblock \emph{arXiv preprint arXiv:2507.02962}, 2025.

\bibitem[Tao et~al.(2025)Tao, Wu, Yin, Zhang, Li, Shen, Li, Zhang, Wang, Jiang, Xie, Huang, and Zhou]{tao2025webshaperagenticallydatasynthesizing}
Zhengwei Tao, Jialong Wu, Wenbiao Yin, Junkai Zhang, Baixuan Li, Haiyang Shen, Kuan Li, Liwen Zhang, Xinyu Wang, Yong Jiang, Pengjun Xie, Fei Huang, and Jingren Zhou.
\newblock Webshaper: Agentically data synthesizing via information-seeking formalization, 2025.
\newblock URL \url{https://arxiv.org/abs/2507.15061}.

\bibitem[Team et~al.(2025)Team, Jaghouar, Mattern, Ong, Straube, Basra, Pazdera, Thaman, Di~Ferrante, Gabriel, et~al.]{team2025intellect}
Prime~Intellect Team, Sami Jaghouar, Justus Mattern, Jack~Min Ong, Jannik Straube, Manveer Basra, Aaron Pazdera, Kushal Thaman, Matthew Di~Ferrante, Felix Gabriel, et~al.
\newblock Intellect-2: A reasoning model trained through globally decentralized reinforcement learning.
\newblock \emph{arXiv preprint arXiv:2505.07291}, 2025.

\bibitem[Trivedi et~al.(2022)Trivedi, Balasubramanian, Khot, and Sabharwal]{trivedi2022musique}
Harsh Trivedi, Niranjan Balasubramanian, Tushar Khot, and Ashish Sabharwal.
\newblock Musique: Multihop questions via single-hop question composition.
\newblock \emph{Transactions of the Association for Computational Linguistics}, 10:\penalty0 539--554, 2022.

\bibitem[Wang et~al.(2024{\natexlab{a}})Wang, Ma, Feng, Zhang, Yang, Zhang, Chen, Tang, Chen, Lin, et~al.]{wang2024survey}
Lei Wang, Chen Ma, Xueyang Feng, Zeyu Zhang, Hao Yang, Jingsen Zhang, Zhiyuan Chen, Jiakai Tang, Xu~Chen, Yankai Lin, et~al.
\newblock A survey on large language model based autonomous agents.
\newblock \emph{Frontiers of Computer Science}, 18\penalty0 (6):\penalty0 186345, 2024{\natexlab{a}}.

\bibitem[Wang et~al.(2024{\natexlab{b}})Wang, Chen, Fu, Liao, Zhang, Wu, Yu, Xu, Zhang, Luo, et~al.]{wang2024leave}
Minzheng Wang, Longze Chen, Cheng Fu, Shengyi Liao, Xinghua Zhang, Bingli Wu, Haiyang Yu, Nan Xu, Lei Zhang, Run Luo, et~al.
\newblock Leave no document behind: Benchmarking long-context llms with extended multi-doc qa.
\newblock \emph{arXiv preprint arXiv:2406.17419}, 2024{\natexlab{b}}.

\bibitem[Wu et~al.(2025{\natexlab{a}})Wu, Li, Fang, Yin, Zhang, Tao, Zhang, Xi, Fu, Jiang, et~al.]{wu2025webdancer}
Jialong Wu, Baixuan Li, Runnan Fang, Wenbiao Yin, Liwen Zhang, Zhengwei Tao, Dingchu Zhang, Zekun Xi, Gang Fu, Yong Jiang, et~al.
\newblock Webdancer: Towards autonomous information seeking agency.
\newblock \emph{arXiv preprint arXiv:2505.22648}, 2025{\natexlab{a}}.

\bibitem[Wu et~al.(2025{\natexlab{b}})Wu, Yin, Jiang, Wang, Xi, Fang, Zhang, He, Zhou, Xie, et~al.]{wu2025webwalker}
Jialong Wu, Wenbiao Yin, Yong Jiang, Zhenglin Wang, Zekun Xi, Runnan Fang, Linhai Zhang, Yulan He, Deyu Zhou, Pengjun Xie, et~al.
\newblock Webwalker: Benchmarking llms in web traversal.
\newblock \emph{arXiv preprint arXiv:2501.07572}, 2025{\natexlab{b}}.

\bibitem[Xbench-Team(2025)]{xbench}
Xbench-Team.
\newblock Xbench-deepsearch, 2025.
\newblock URL \url{https://xbench.org/agi/aisearch}.

\bibitem[Xi et~al.(2025)Xi, Chen, Guo, He, Ding, Hong, Zhang, Wang, Jin, Zhou, et~al.]{xi2025rise}
Zhiheng Xi, Wenxiang Chen, Xin Guo, Wei He, Yiwen Ding, Boyang Hong, Ming Zhang, Junzhe Wang, Senjie Jin, Enyu Zhou, et~al.
\newblock The rise and potential of large language model based agents: A survey.
\newblock \emph{Science China Information Sciences}, 68\penalty0 (2):\penalty0 121101, 2025.

\bibitem[Xu et~al.(2024)Xu, Fu, Gao, Ye, Liu, Mei, Wang, Yu, and Wu]{xu2024dposuperiorppollm}
Shusheng Xu, Wei Fu, Jiaxuan Gao, Wenjie Ye, Weilin Liu, Zhiyu Mei, Guangju Wang, Chao Yu, and Yi~Wu.
\newblock Is dpo superior to ppo for llm alignment? a comprehensive study, 2024.
\newblock URL \url{https://arxiv.org/abs/2404.10719}.

\bibitem[Yang et~al.(2018)Yang, Qi, Zhang, Bengio, Cohen, Salakhutdinov, and Manning]{yang2018hotpotqa}
Zhilin Yang, Peng Qi, Saizheng Zhang, Yoshua Bengio, William~W Cohen, Ruslan Salakhutdinov, and Christopher~D Manning.
\newblock Hotpotqa: A dataset for diverse, explainable multi-hop question answering.
\newblock \emph{arXiv preprint arXiv:1809.09600}, 2018.

\bibitem[Yao et~al.(2023)Yao, Zhao, Yu, Du, Shafran, Narasimhan, and Cao]{yao2023react}
Shunyu Yao, Jeffrey Zhao, Dian Yu, Nan Du, Izhak Shafran, Karthik Narasimhan, and Yuan Cao.
\newblock React: Synergizing reasoning and acting in language models.
\newblock In \emph{International Conference on Learning Representations (ICLR)}, 2023.

\bibitem[Yu et~al.(2025)Yu, Zhang, Zhu, Yuan, Zuo, Yue, Dai, Fan, Liu, Liu, et~al.]{yu2025dapo}
Qiying Yu, Zheng Zhang, Ruofei Zhu, Yufeng Yuan, Xiaochen Zuo, Yu~Yue, Weinan Dai, Tiantian Fan, Gaohong Liu, Lingjun Liu, et~al.
\newblock Dapo: An open-source llm reinforcement learning system at scale.
\newblock \emph{arXiv preprint arXiv:2503.14476}, 2025.

\bibitem[Yu et~al.(2024)Yu, Zhang, and Feng]{yu2024auto}
Tian Yu, Shaolei Zhang, and Yang Feng.
\newblock Auto-rag: Autonomous retrieval-augmented generation for large language models.
\newblock \emph{arXiv preprint arXiv:2411.19443}, 2024.

\bibitem[Zhao et~al.(2025)Zhao, Gao, Song, Chen, Yang, Fu, Wang, Iwasawa, Matsuo, and Li]{zhao2025reagent}
Xinjie Zhao, Fan Gao, Xingyu Song, Yingjian Chen, Rui Yang, Yanran Fu, Yuyang Wang, Yusuke Iwasawa, Yutaka Matsuo, and Irene Li.
\newblock Reagent: Reversible multi-agent reasoning for knowledge-enhanced multi-hop qa.
\newblock \emph{arXiv preprint arXiv:2503.06951}, 2025.

\bibitem[Zheng et~al.(2025)Zheng, Fu, Hu, Cai, Ye, Lu, and Liu]{zheng2025deepresearcher}
Yuxiang Zheng, Dayuan Fu, Xiangkun Hu, Xiaojie Cai, Lyumanshan Ye, Pengrui Lu, and Pengfei Liu.
\newblock Deepresearcher: Scaling deep research via reinforcement learning in real-world environments.
\newblock \emph{arXiv preprint arXiv:2504.03160}, 2025.

\end{thebibliography}
}

\appendix
\section{Full Case Study}
\label{app:case-study}

\begin{figure}
    \centering
    \includegraphics[width=0.9\linewidth]{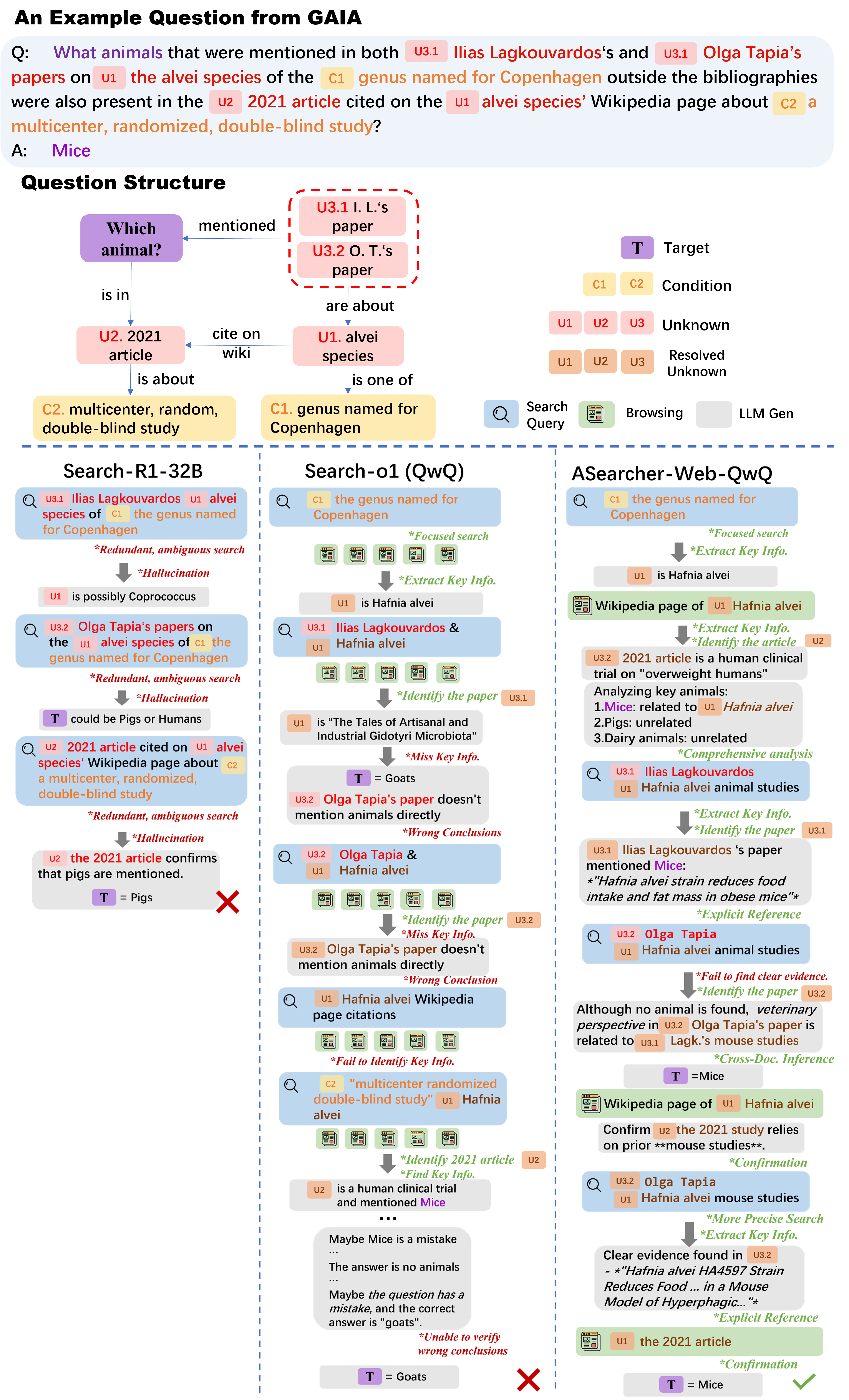}
    \caption{A case study on a complex query from GAIA. \textbf{Search-R1-32B} is unable to break down the complex question and has severe hallucinations. \textbf{Search-o1 (QwQ)} can identify the corrects articles through extensive tool calls, but easily misses key information and fails to verify wrong conclusions. Our end-to-end RL agent, \textbf{{\name}-Web-QwQ}, exhibits key behaviors featuring Search Intelligence: \emph{uncertainty-aware reasoning} (list and examine candidate answers), \emph{precise extraction} from noisy contents, \emph{cross-document inference}, and \emph{rigorous confirmation}.}
    \label{fig:case-study}
\end{figure}

In this section, we provide a detailed case study on an extremely challenging question from GAIA~\citep{mialon2023gaia}. Specifically, we analyze Search-R1-32B~\citep{jin2025search} and Search-o1 (QwQ)~\citep{li2025search} in Fig.~\ref{fig:case-study}.

\paragraph{Solution Path of the Sample Question.} In Fig.~\ref{fig:case-study}, our case study is carried out on a question requiring finding some specific animal given 2 \textbf{\textcolor{orange}{conditions}} and 4 \textbf{\textcolor{red}{unknown variables}}. To identify the correct answer, the search agent should first find out the mentioned species \textbf{\textcolor{red}{U1}} according to condition \textbf{\textcolor{orange}{C1}}, identify the correct article \textbf{\textcolor{red}{U2}} that satisfies condition \textbf{\textcolor{orange}{C2}}, and then find out the papers listed in \textbf{\textcolor{red}{U3.1}} and \textbf{\textcolor{red}{U3.2}}. Finally, the correct answer should be determined by cross referencing the article \textbf{\textcolor{red}{U2}} and the papers \textbf{\textcolor{red}{U3.1}}\&\textbf{\textcolor{red}{U3.2}}. To summarize, this example is challenging for several main reasons,
\begin{itemize}
\item \textbf{High Uncertainty:} The question involves multiple unknown variables that could point to many different entities. For example, the 2021 article \textbf{\textcolor{red}{U2}} could point to any article published in 2021 and could only be determined given the condition \textbf{\textcolor{orange}{C2}} and the alvei species \textbf{\textcolor{red}{U1}}. 
\item \textbf{Requirement for Exact Information Extraction:} To find the answer, the agent should list all animals mentioned on the webpages and making cross-document comparison. This would require the agent to precisely extract key information from the vast, noisy web contents, instead of simply summarizing the webpages.
\item \textbf{Misleading Answers:} During the process of solving this task, there could be multiple misleading answers, such as "pigs". The agent should rigorously confirm its conclusions by checking the intended answer in all related webpages and documents.
\end{itemize}

\paragraph{Existing Online RL Approaches Fail to Learn Complex Search Strategies.} 
In Fig.~\ref{fig:case-study}, Search-R1-32B is not able to decompose the complex query into individual components, consequently only making redundant queries that involve too many unknown information. The agent also has severe hallucinations, producing conclusions that are not supported by the search results. Finally, it fails to resolve all unknown variables. This case study shows that existing online RL approaches only incentivize elementary search strategies. It is also worth noting that, since the turn limit is set as a small value, e.g. 4, during training, the model only exhibits a short tool-use horizon.

\paragraph{Prompt-based LLM Agents Could Fail Due to Insufficient Capability of the LLM.} In Fig.~\ref{fig:case-study}, Search-o1 (QwQ) can find the species name \textbf{\textcolor{red}{U1}}, as well as the 2021 article \textbf{\textcolor{red}{U2}} and papers \textbf{\textcolor{red}{U3.1\&U3.2}} through a large amount of tool calls. However, when trying to find the answer, Search-o1 (QwQ) would easily miss key information. Consequently, the agent makes incorrect conclusions. Notably, even when the agent finds information that directly links to the correct answer, it is still misguided by previous incorrect conclusions. Finally, the agent is unable to verify the correctness of previous conclusions. This case study reveals that, though an open-source model that is not explicitly trained on agentic tasks can perform extensive tool calls, it could not make expert-level reasoning based on the retrieved contents and history contexts.

\paragraph{{\name}-Web-QwQ.} We also analyze the search strategy of our end-to-end RL agent, {\name}-Web-QwQ.
As shown in Fig.~\ref{fig:case-study}, {\name}-Web-QwQ decomposes the complex query into precise and focused queries. Unlike Search-o1 (QwQ) that visits a large amount of websites after each search query, {\name}-Web-QwQ focuses on visiting the most relevant website. {\name}-Web-QwQ summarizes all related information from a website. Specifically, all candidate answers are listed and carefully analyzed by the agent. When trying to search for related facts in the papers \textbf{\textcolor{red}{U3.1\&U3.2}}, the agent explicitly references the key information. When the search results do not directly point to the desired target, e.g. when searching with ``\textcolor{red}{Olga Tapia (\textbf{U3.2})} \textcolor{brown}{Hafnia alvei (\textbf{U1})} animal studies'' to find the animals related to Olga Tapia's paper, the agent does not get a clear information but is able to infer the correct answer by make connection with the other paper \textcolor{red}{\textbf{U3.1}}. After the correct answer ``Mice'' is found, the agent spends further turns on confirming previous conclusions before reporting the final answer. In summary, {\name} successfully train a search agent that exhibits complex behaviors that feature Search Intelligence, 
\begin{itemize}
\item \textbf{Uncertainty-aware reasoning:} the agent exhaustively lists and examines all possibilities for uncertain entities
\item \textbf{Price Key Information Extraction:}  the agent is able to identify the key information from vast, noisy web contents.
\item \textbf{Cross-document Inference:} the agent is able to infer critical conclusions by making connections among multiple documents.
\item \textbf{Rigorous Confirmation:} the agent verifies the correctness of previous conclusions with additional tool calls. 
\end{itemize}

\end{document}